\documentclass[10pt,twocolumn,letterpaper]{article}

\usepackage{cvpr}              

\usepackage[dvipsnames]{xcolor}

\usepackage{amsmath}
\usepackage{url}
\definecolor{cvprblue}{rgb}{0.21,0.49,0.74}
\usepackage[pagebackref,breaklinks,colorlinks,citecolor=cvprblue]{hyperref}
\usepackage[accsupp]{axessibility} 
\usepackage{multirow}
\usepackage{makecell}

\usepackage{algorithm}

\usepackage{algpseudocode}

\def\paperID{12109} 
\def\confName{CVPR}
\def\confYear{2024}

\title{
\includegraphics[width=0.044\textwidth]{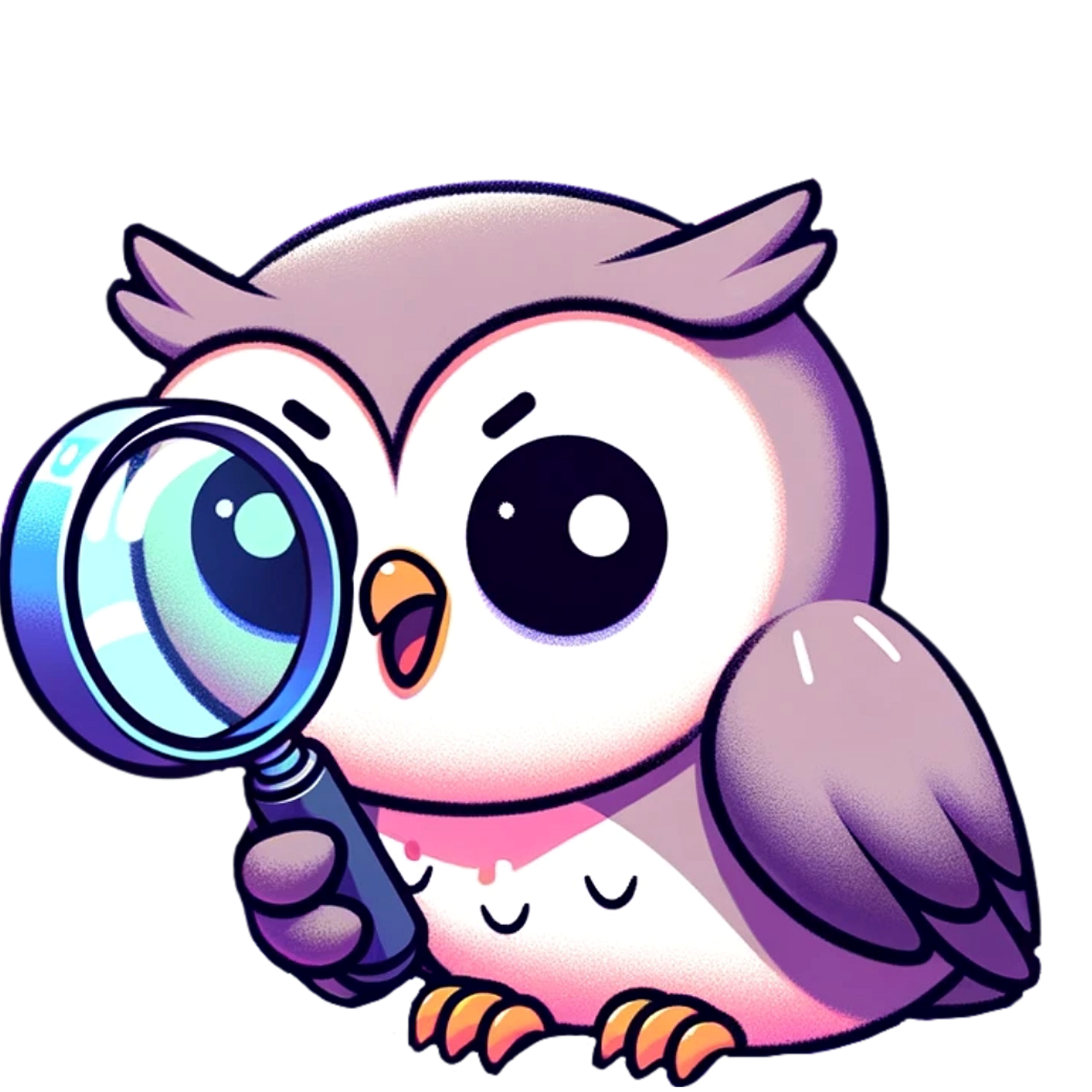} Uncovering What, Why and How: \\
 A Comprehensive Benchmark for Causation Understanding of Video Anomaly  
 }

\author{
    Hang Du\textsuperscript{\rm 1}\footnotemark[1],
    Sicheng Zhang\textsuperscript{\rm 2}\footnotemark[1],
    Binzhu Xie\textsuperscript{\rm 2}\footnotemark[1],
    Guoshun Nan\textsuperscript{\rm 1}\footnotemark[2],
    Jiayang Zhang\textsuperscript{\rm 1},
    Junrui Xu\textsuperscript{\rm 1},
    Hangyu Liu\textsuperscript{\rm 1},\\
    Sicong Leng\textsuperscript{\rm 3}, 
    Jiangming Liu\textsuperscript{\rm 4},
    Hehe Fan\textsuperscript{\rm 5},
    Dajiu Huang\textsuperscript{\rm 6},
    Jing Feng\textsuperscript{\rm 6},
    Linli Chen\textsuperscript{\rm 6},
    Can Zhang\textsuperscript{\rm 1},\\
    Xuhuan Li\textsuperscript{\rm 1},
    Hao Zhang\textsuperscript{\rm 1}, 
    Jianhang Chen\textsuperscript{\rm 1},
    Qimei Cui\textsuperscript{\rm 1},
    Xiaofeng Tao\textsuperscript{\rm 1}
    \and  
  \textsuperscript{\rm 1}Beijing University of Posts and Telecommunications
  \textsuperscript{\rm 2}Queen Mary University of London \and 
  \textsuperscript{\rm 3}Nanyang Technological University 
  \textsuperscript{\rm 4}Yunnan University \and 
  \textsuperscript{\rm 5}Zhejiang University     
  \textsuperscript{\rm 6}China Telecom Co., Ltd. Sichuan Branch
    }

\begin{document}


\twocolumn[{%
   \renewcommand\twocolumn[1][]{#1}%
   \maketitle
   \vspace{-25pt}
   \begin{center}
    \centering
    \includegraphics[width=1.0\linewidth]{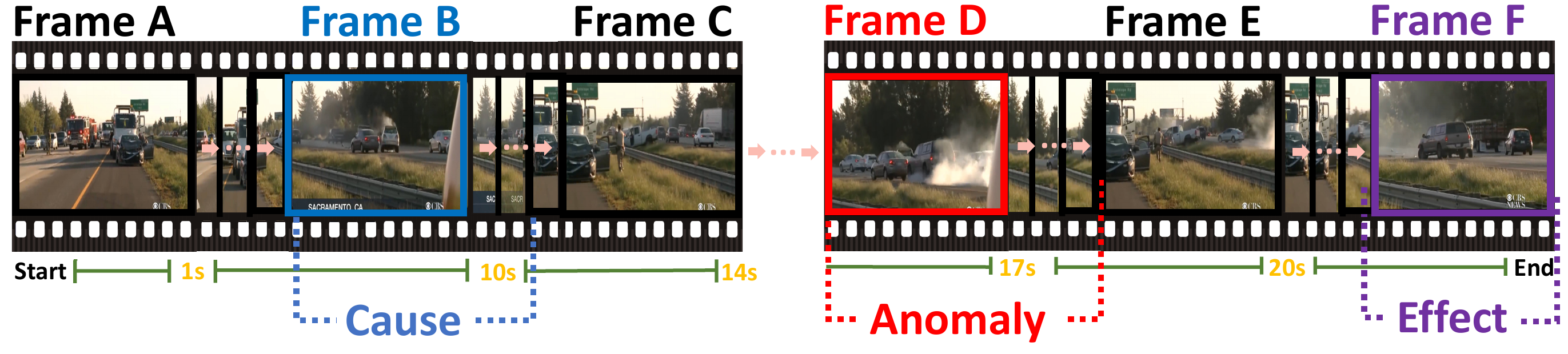}
    \vspace{-10pt}    
    \captionof{figure}{\textbf{Illustration of causations of video anomaly.} The clip started at Frame D refers to a traffic accident, which was caused by the event indicated with Frame B 7 seconds before. The clip in Frame F shows the effect of such an anomaly. A model needs to understand such a long-range relation in the video to yield correct text-based explanations.}
    \vspace{-5pt}
    \label{fig:intro}
   \end{center}%
  }]
\renewcommand{\thefootnote}{\fnsymbol{footnote}} 
\footnotetext[1]{Equal Contribution}
\footnotetext[2]{Corresponding Author: Guoshun Nan (nanguo2021@bupt.edu.cn)}
\begin{abstract}
Video anomaly understanding (VAU) aims to automatically comprehend unusual occurrences in videos, thereby enabling various applications such as traffic surveillance and industrial manufacturing. While existing VAU benchmarks primarily concentrate on anomaly detection and localization, our focus is on more practicality, prompting us to raise the following crucial questions: ``what anomaly occurred?'', ``why did it happen?'', and ``how severe is this abnormal event?''. In pursuit of these answers, we present a comprehensive benchmark for Causation Understanding of Video Anomaly (CUVA). Specifically, each instance of the proposed benchmark involves three sets of human annotations to indicate the ``what'', ``why'' and ``how'' of an anomaly, including 1) anomaly type, start and end times, and event descriptions, 2) natural language explanations for the cause of an anomaly, and 3) free text reflecting the effect of the abnormality. In addition, we also introduce MMEval, a novel evaluation metric designed to better align with human preferences for CUVA, facilitating the measurement of existing LLMs in comprehending the underlying cause and corresponding effect of video anomalies. Finally, we propose a novel prompt-based method that can serve as a baseline approach for the challenging CUVA. We conduct extensive experiments to show the superiority of our evaluation metric and the prompt-based approach.
Our code and dataset are available at \url{https://github.com/fesvhtr/CUVA}.
\end{abstract}    
\section{Introduction}


\par Anomalies represent occurrences or scenarios that deviate from the norm, defying expectations and straying from routine conditions \cite{9151050, acsintoae2022ubnormal,VideoAnomaly,adam2008robust}. These events are typically characterized by their unique, sudden, or infrequent nature, often demanding special attention or intervention \cite{ramachandra2020street}. 

Recently proliferated video anomaly understanding (VAU) \cite{lv2023unbiased,wu2022self} aims at automatically comprehending such abnormal events in videos, thereby facilitating various applications such as traffic surveillance,  environmental monitoring, and industrial manufacturing. In this direction, video anomaly detection and localization, which refer to identifying abnormal occurrences, and localizing temporally or spatially locate anomalous events in videos, has attracted enormous attention \cite{mehran2009abnormal,wu,velastin2017people,xu2022tad,de2022camnuvem,Wang2023,10083244,Birds,10203301}.
  
Existing VAU benchmarks \cite{cao2023new,han2022adbench,thakare2023rareanom} and approaches \cite{zaheer2022generative,singh2023eval,chang2022video,nan1,nan2,shi1,shi2,liujun1, liujun2} primarily focus on the aforementioned anomaly detection and localization tasks, while the underlying cause and the corresponding effect of these occurrences, are still largely under-explored. These cues are crucial for perceiving the abnormality and making decisions based on human-interpretable explanations. Figure \ref{fig:intro}
demonstrates a scene of a traffic accident involving many vehicles. ``The accident occurred because a white car parked by the roadside, and a dark gray car traveled at high speed to swerve and rear-end the black car next to it.'' Challenges of comprehending such a cause of the accident include: 1) \textit{capturing key cues in the long video:} a model needs to recognize the white car at the moment indicated by Frame B, which is $7$ seconds before the accident in the clip indicated by Frame D. It is challenging for a model to capture such a long-range relation. 2) \textit{building a logic chain of the cause-effect:} a model needs to further learn rich interactions among clips in the video, indicated by Frame B, Frame C, and Frame D, to build a logic chain of causation of the anomaly, facilitating the generation of the explanations and results. The above two challenges require the development of causation understanding methods that specifically take these characteristics of video anomaly into consideration.

Previous works have demonstrated the great importance of leveraging large, high-quality, and challenging benchmarks to develop and evaluate the state-of-the-art deep learning methods for the VAU task \cite{anomalyde, liu2023generating, aboah2021vision, tian2021weakly, nguyen2019anomaly, ruff2019deep}. 
Along this line, existing benchmarks have shown their promise \cite{wu, VideoAnomaly, ramachandra2020street}. Towards VAU in more practical real-world scenarios, they have some limitations:
1) \textit{Lack of cause and effect explanations.} Existing annotations involve the periods when anomalies occur, without providing an explanation of the underlying cause and the effect, as well as the descriptions of targeting anomaly.  
2) \textit{Lack of proper evaluation metrics.} Some remotely related metrics to measure the text-based explanation or description of the video anomaly, such as BLEU \cite{papineni2002bleu} and ROUGE \cite{lin2004rouge}, can not be directly applied to measure multimodal VAU tasks, as they are designed only for text modality.  
3) \textit{Limited length of videos.} In real-world scenarios, a piece of video may include more than 1.5 minutes \cite{apostolidis2021video}. However, samples in existing VAU usually have fewer than 30 seconds, which greatly simplifies the challenges of VAU in real-world cases.

The above limitations of existing datasets call for a benchmark of Causation Understanding of Video Anomaly.
Towards that, we present CUVA, a comprehensive benchmark that contains high-quality annotations of $1,000$ videos from the real world, covering $10$ major categories, and $42$ subcategories of different anomaly types, each involving a 117-second long video and ``65.7'' tokens across ``4.3'' sentences on average.
Specifically, we manually write free-text explanations to detail the underlying cause and the corresponding effects, the descriptions of these events, and the relationships among them. 
Moreover, we come up with a novel evaluation metric to measure the capability of a method on the challenging CUVA. We also propose a novel prompt-based approach based on video large language model (VLM) \cite{video-chatgpt, mplug-owl,2023videochat}. Experiments show the superiority of the metric and the proposed method. The main contributions of our work can be summarised as follows:

\begin{itemize}
	\item We develop CUVA, a new benchmark for causation understanding of video anomaly. To the best of our knowledge, CUVA is the first large-scale benchmark focused on the causation of video anomalies. Compared with existing datasets, our dataset is more comprehensive and more challenging with much higher-quality annotations.
	\item We present a novel metric to measure the challenging CUVA in a human-interpretable manner, and introduce a prompt-based method to capture the key cues of anomalies and build a logic chain of the cause-effect. 
	\item We conduct extensive experiments on the proposed CUVA. Results show that CUVA enables us to develop and evaluate various VLM methods for causation understanding of video anomalies closer to real-world cases.
\end{itemize}

\section{Related Work}
\label{sec: Related Work}
\begin{figure*}[h!]
    \vspace{0pt}
    \centering
    \includegraphics[width=1\linewidth]{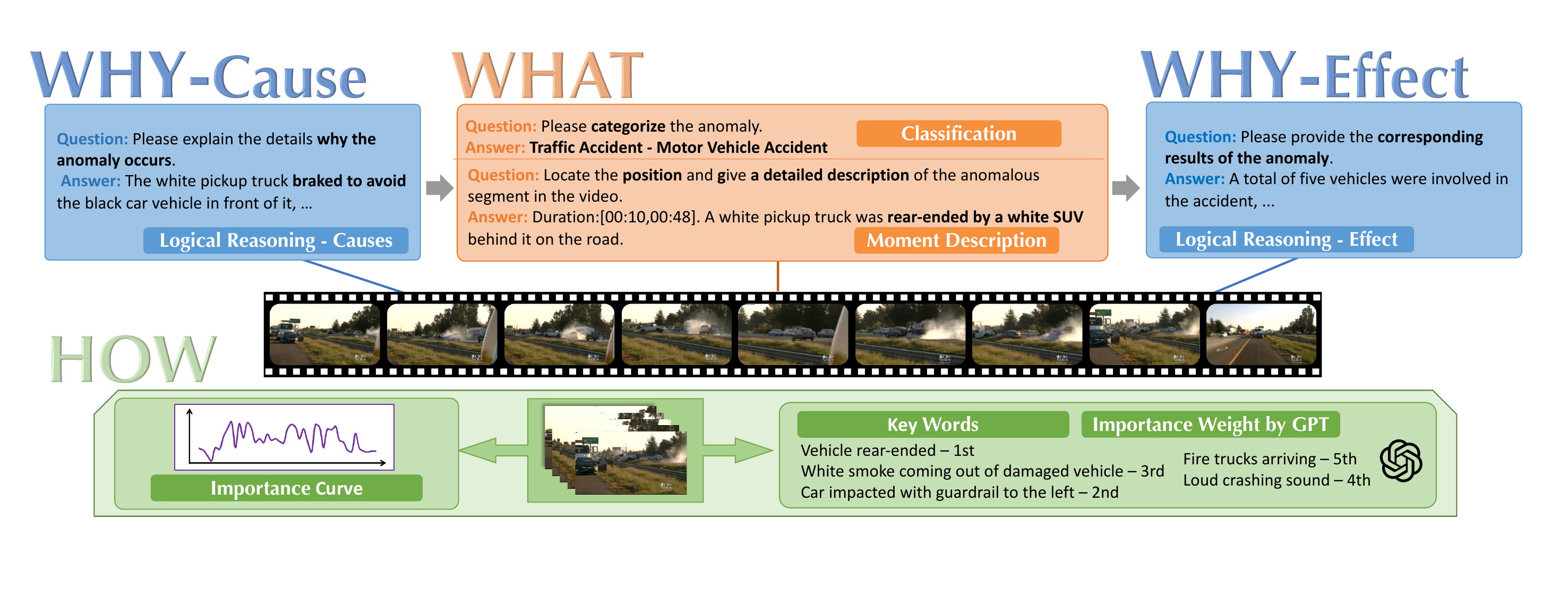}
    \captionof{figure}{
    \textbf{Overview of the proposed CUVA benchmark.} Our CUVA benchmark consists of manual text-based annotation, including detailed explanations of cause (Why) and effect (Why), anomaly types (What), detailed event descriptions (What), as well as importance scores that can form a curve of events (How).}
    \vspace{0cm}
    \label{fig:dataset}
\end{figure*}


\noindent
\textbf{Anomaly Datasets:} 
Existing VAU datasets primarily focus on anomaly detection and localization, and can be broadly categorized into weakly-supervised ones \cite{sultani2018real,wu}, and semi-supervised ones \cite{luo2017revisit,acsintoae2022ubnormal,ramachandra2020street,rodrigues2020multi}.
These datasets emphasize the time points or time periods of anomalous events based on frame-level or pixel-level annotations.
Our CUVA significantly differs from the existing datasets in these aspects, 
More detailed comparisons are available in Table \ref{comparison}.


\noindent
\textbf{Evaluation Metrics:} 
VAU evaluation metrics \cite{xu2023critical} include,
reference-based ones such as ROUGE \cite{lin2004rouge} and BLEURT \cite{sellam2020bleurt}, answer-based ones such as BLEU \cite{papineni2002bleu}, Rankgen \cite{krishna2022rankgen} and QAFactEval \cite{fabbri2021qafacteval}, and others such as Longformer \cite{beltagy2020longformer}, UniEval \cite{zhong-etal-2022-towards} and MoverScore \cite{zhao-etal-2019-moverscore}.
Recently, various GPT-based metrics \cite{xie2023funqa,touchstone,VisITBench} have been developed. 
The key difference between our proposed MMEval and the above ones is: MMEval aims to evaluate the video and text anomaly understanding based on a large language model, while the existing one focuses on a single modality.



\noindent
\textbf{Methods:} 
Video large language models (VLM) have been widely used for text generation based on videos \cite{li2023otter,su2023pandagpt,damonlpsg2023videollama,video-chatgpt}, exploring
prompts to unlock the capability of VLMs. Prompt-based methods can be categorized into ``hard prompt'' and ``soft prompt'' \cite{2022arXiv221006466L,2021arXiv211201518R,ding2021openprompt,2021arXiv210613353L}.
For the challenging CUVA task, we proposed a novel method that leverages both hard prompts and soft ones to tackle two challenges raised at the beginning, i.e., capturing the key cues and building a logic chain of anomaly causation.


\section{The Proposed CUVA Benchmark} \label{sec:Dataset}
In this section, we first introduce some CUVA sub-tasks. Then we show how we collect and annotate data.
We also provide a quantitative analysis of the benchmark. The overview of our CUVA is demonstrated in Figure \ref{fig:dataset}.

\subsection{Task Definition}
\begin{figure}[h]
    \centering
    \includegraphics[width=0.45\textwidth]{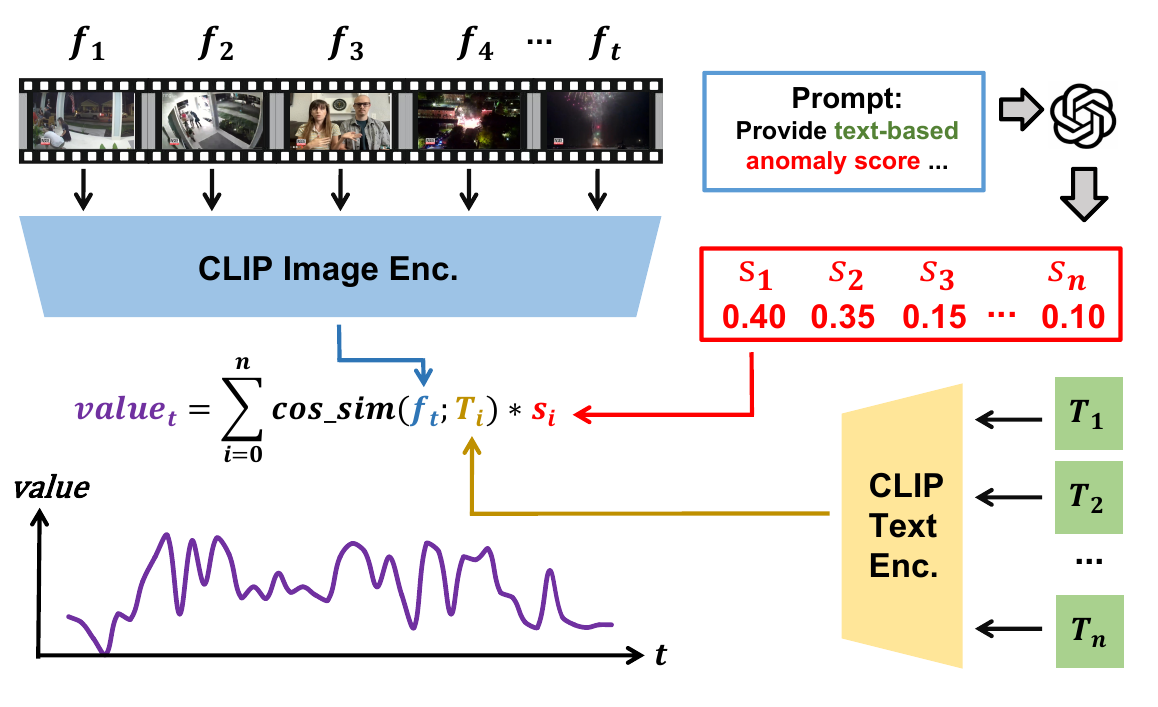}
    \caption{\textbf{Pipeline of generating an importance curve.} Annotators need to consider previous tasks (e.g., Logical Description, Moment Description) and video content to create $3$ to $6$ short sentences ${T_{i}}$ describing all events in the video. We rank these sentences' anomaly severity by ChatGPT \cite{chatgpt} and obtain anomaly scores $s$. Simultaneously, we sample frames ${f_{t}}$ from the video and use CLIP \cite{clip} to measure the similarity between sentences and frames. The resulting similarity scores are multiplied by the anomaly scores for each sentence to get $value_{t}$ for each frame.}
    \label{fig:pipline}
    \vspace{-3mm}
\end{figure}

\textbf{What anomaly occurred}: This task includes two objectives: anomaly classification and anomaly description. \textit{Anomaly Classification} includes all the anomaly classes present in the video, which are taken from our database of predefined anomaly classes, as shown in Figure \ref{fig:statics} (a).  
Each video has multiple anomaly classes at different levels, and this task will challenge the model's ability to detect anomaly classes at multiple levels of granularity. \textit{Anomaly Moment Description} includes the timestamp in which the anomaly occurs and a detailed description of the anomalous event. 
\newline\textbf{Why this anomaly happened}: This task aims to describe the causal relationships within the video.
Anomaly reasoning describes the reasons for the occurrence of anomalies in the video. 
This task requires the model to infer the cause of the anomaly based on the video content and describe it in natural language, which challenges the model's ability of video comprehension and reasoning.
Anomaly results primarily describe the impacts caused by anomalous events in the video. It mainly tests the model's ability to handle details of anomalous events in the video.
\newline\textbf{How severe this anomaly}: This task aims to reflect the changing trends in the severity of anomalies within the video.
Thus, we propose a novel annotation approach called the importance curve.
Details of our importance curve's pipeline can be found in Figure \ref{fig:pipline}. 
This approach has three advantages: 
1) It provides an intuitive representation of the temporal variation in anomaly severity within the video. 
2) It offers a more intuitive depiction of the inherent causal relationships among anomalous events in the video. 
3) Such an approach enables us to unify various Video Temporal Grounding labels and tasks (e.g. Moment Retrieval, Highlight Detection, Video Summarization) under the same framework.\footnote{More details are available in Section 2 of Appendix A.}

\subsection{Dataset Collection}
We crawled data from prominent video platforms such as Bilibili and YouTube\footnote{We have obtained permission from Bilibili \url{www.bilibili.com} and YouTube \url{www.youtube.com} to use their video data for non-commercial purposes.}. 
And we discarded videos that encompass sensitive themes such as pornography and politics. 
Throughout the data collection process, we thoroughly analyze the quantity and quality of videos in each category, which in turn lead to the selection of the final $11$ categories of anomalous videos. 
These videos are then categorized into $11$ main categories, such as ``robbery", ``traffic accident" and ``fire".  
Each major category is further divided into subcategories. 
For example,  we divided the ``fire'' category into the ``commercial building fire'', ``forest fire'', ``factory fire'' and ``residential fire'' subcategories. 
In this way, we obtain $42$ subcategories in total.  
\begin{table*}[!h]
\resizebox{1.0\linewidth}{!}{
\begin{tabular}{lcccccccccc}
\toprule
\multirow{2}{*}{Dataset} & \multirow{2}{*}{Domain} & \multicolumn{4}{c}{Video} & \multirow{2}{*}{\# Anomaly Types} & \multicolumn{4}{c}{QA}  \\
\cmidrule(lr){3-6} \cmidrule(lr){8-11}
 &  &  \# Total Frames & Total Length & A.C.L & Audio & & Localization & Description & Reasoning & Outcome  \\
\midrule

UCF-Crimes \cite{sultani2018real} & Crime & 13,741,393 & 128.0h& 242.5s & No & 13 & Frame & NA & NA & NA \\
XD-Violence \cite{wu} & Volence  & 114,096 & 21.07h & 164.3s & Yes & 6 & Frame & NA & NA & NA \\
ShanghaiTech \cite{luo2017revisit}  & Pedestrian & 317,398 & - & - & No & 13 & Bounding-box & NA & NA & NA \\
UCSD Ped1 \cite{wang2010anomaly} & Pedestrian & 14,000 & 0.1h & 6.6s & No & 5 & Bounding-box& NA & NA & NA \\
UCSD Ped2 \cite{wang2010anomaly} & Pedestrian & 4,560 & 0.1h & 6.6s & No & 5 & Bounding-box & NA & NA & NA \\
CUHK Avenue \cite{abnormal2013lu} & Pedestrian & 30,652 & 0.5h & 1.4s & No & 5 & Bounding-box & NA & NA & NA \\
TAD \cite{xu2022tad} & Trafﬁc & 721,280 & 1.2h & 36.8s & Irrelevant & 4 & Bounding-box & NA & NA & NA \\
Street Scene \cite{ramachandra2020street} & Trafﬁc & 203,257 & 380.6s & 3.7s & No & 17 & Bounding-box & NA & NA & NA \\
CamNuvem \cite{de2022camnuvem} & Robbery & 6,151,788 & 57h & 192.2s & No & 1 & Frame & NA & NA & NA \\
Subway Entrance \cite{adam2008robust} & Pedestrian &  86,535 & 1.5h & - & No & 5 & Frame & NA & NA & NA \\
Subway Exit \cite{adam2008robust} & Pedestrian & 38,940 & 1.5h & - & No & 3 & Frame & NA & NA & NA \\
UCF–Crime Extension \cite{ozturk2021adnet} & Crime &  734,400 & 7.5h & 112.5s & No & 1 & Frame & NA & NA & NA \\
BOSS \cite{velastin2017people} & Multiple & 48,624 & 0.5h & 660.0 s & No & 11 & Frame & NA & NA & NA \\
UMN \cite{mehran2009abnormal} & behaviors & 3,855 & 0.1h & 29.1s & No & 1 & Frame & NA & NA & NA \\
UBnormal \cite{acsintoae2022ubnormal} & Multiple & 236,902 & 2.2h & 14.6s & No & 22 & Pixel-level & NA & NA & NA \\
\textbf{CUVA (Ours)} & Multiple & 3,345,097 & 32.5h & 117.0s & Yes & \textbf{42} & \textbf{Time Duration} & \textbf{Free-text} & \textbf{Free-text} & \textbf{Free-text} \\ 

\bottomrule
\end{tabular}}
\caption{\textbf{Comparisons between the proposed CUVA and existing VAU datasets.}
Our CUVA is the first large-scale benchmark for causation understanding of video anomaly. It encompasses samples from 42 domains, such as vandalism, traffic accidents, fire incidents, and pedestrian incidents, etc. CUVA sub-tasks primarily focus on the evaluation of causation understanding of video anomaly, and these tasks answer the ``What'', ``Why'' and ``How'' of an anomaly. All textual descriptions or explanations are annotated in \textbf{free-text} format. Here
\textbf{A.C.L.} typically stands for ``Average Clip Length.''
}
\label{comparison}

\end{table*}
\subsection{Annotation Pipeline}
\label{section 3.3 ano pip}
Our dataset construction pipeline involves three stages: pre-processing, manual annotation, and importance curve processing. 
The whole process takes about $150$ hours with over $20$ annotators.\footnote{More details of our dataset are available in Section 3 of Appendix A.}
\subsubsection{Pre-processing}
First, we crawl videos from Bilibili and YouTube. Then, we manually cut the collected videos to ensure the quality of video content and exclude non-ethical content and sensitive information through manual screening.\footnote{Detailed screening criteria can be found in Section 4 of Appendix A.} Throughout the dataset collection and annotation process, we strictly follow the ethical requirement of the website.\footnote{More details about ethical consideration are presented in Section 5 of Appendix A.} Finally, $1,000$ anomaly video clips are obtained.\\
\subsubsection{Manual Annotation}
We annotate the videos in English according to the designed annotation document, and the annotation is divided into two rounds.
We employ a mechanism similar to kappa \cite{XIA2020309} to screen and train annotators, ensuring the consistency of their annotation content.
In the first round, We ask annotators to annotate all videos according to the task definition. In the second round, we ask these annotators to review and supplement the annotation results of the first round.
\subsubsection{Post-processing of Importance Curve}
Due to the limited capabilities of the CLIP model and sampling intervals, the initial curve may fail to accurately reflect the time periods of anomalies, which significantly impacts the effectiveness of downstream tasks. 
Thus, we incorporate the following three tasks to optimize the importance curve, such as Video Captioning \cite{2023videochat}, Video Entailment \cite{sevila}, and Video Grounding \cite{univtg} respectively. 
Based on these tasks, we employ a voting mechanism to precisely identify the time segments in the video corresponding to the given key sentences.\footnote{Details can be found in Section 6 of Appendix A.}



\begin{figure*}[!t]
    \centering
    \begin{subfigure}[b]{0.4\textwidth}
        \centering
        \hspace*{-40pt}
        \includegraphics[width=0.8\linewidth]
        {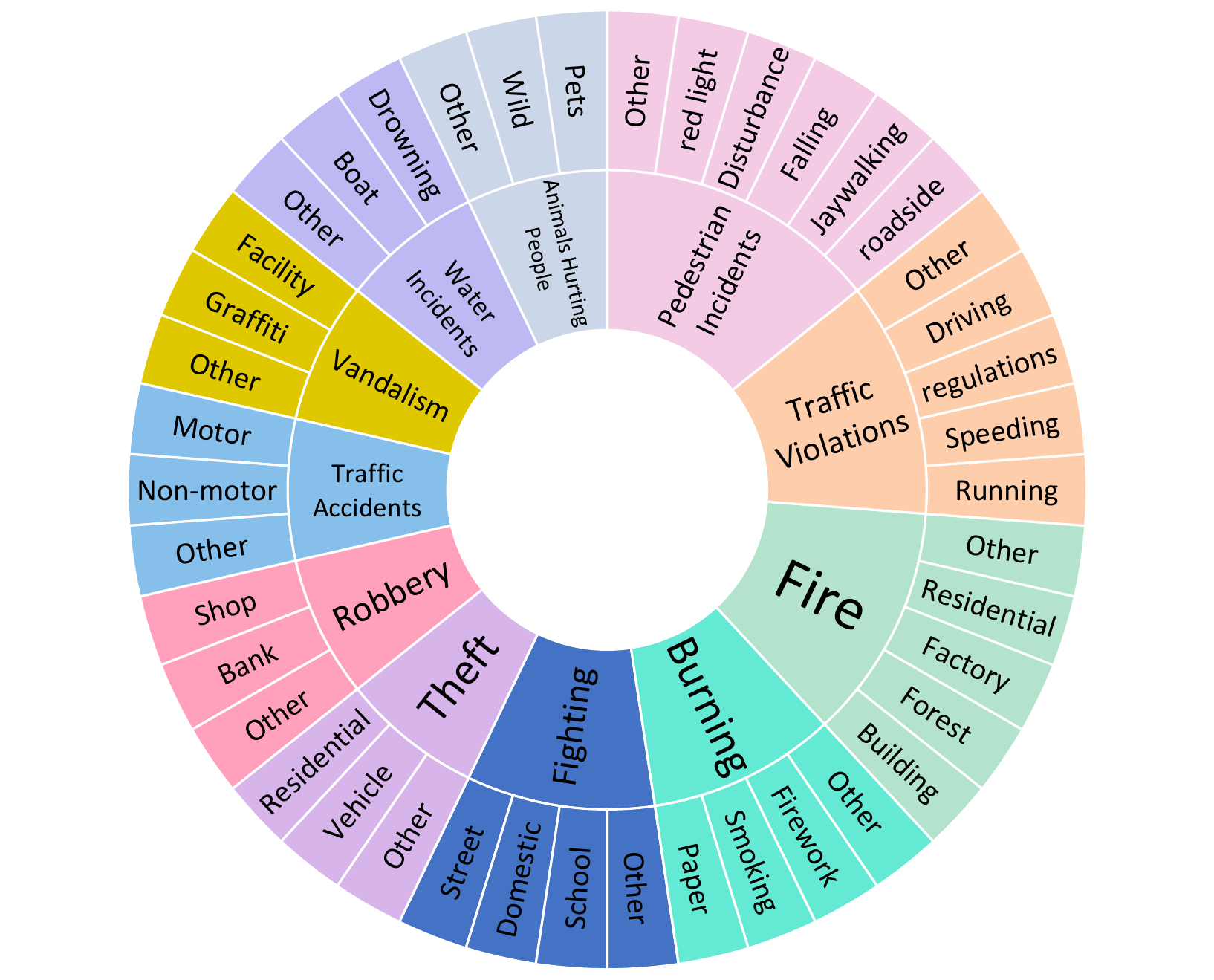}
        \hspace*{-40pt}
        \caption{}
        \label{fig:classes}
    \end{subfigure}
    \begin{subfigure}[b]{0.4\textwidth}
        \centering
        \hspace*{-30pt}
        \includegraphics[width=1.1\linewidth]{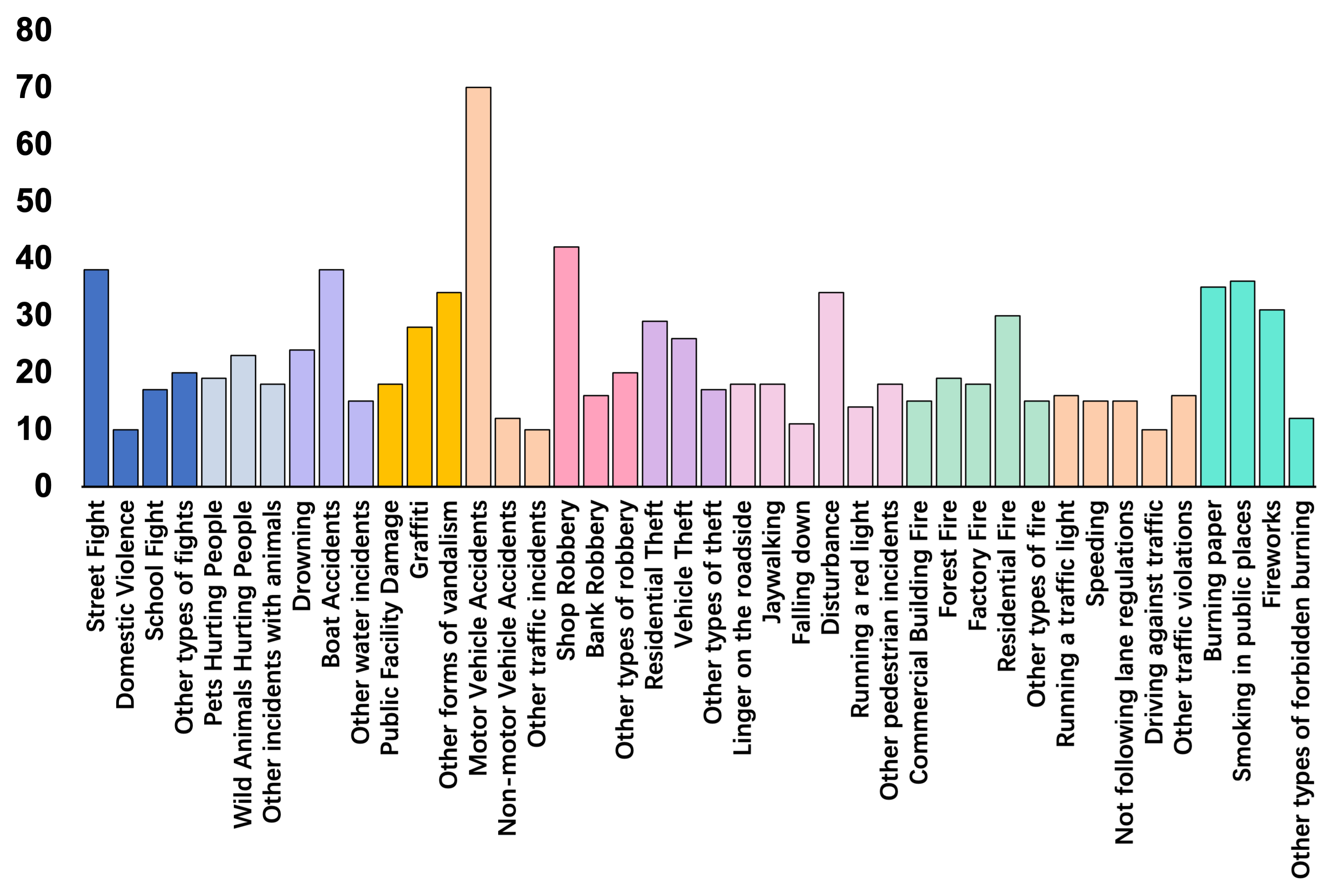}
        \hspace*{-30pt}
        \caption{}
        \label{fig:anomaly_2nd}
    \end{subfigure}

    \begin{subfigure}[b]{0.3\textwidth}
        \centering
        \includegraphics[width=0.8\linewidth]
        {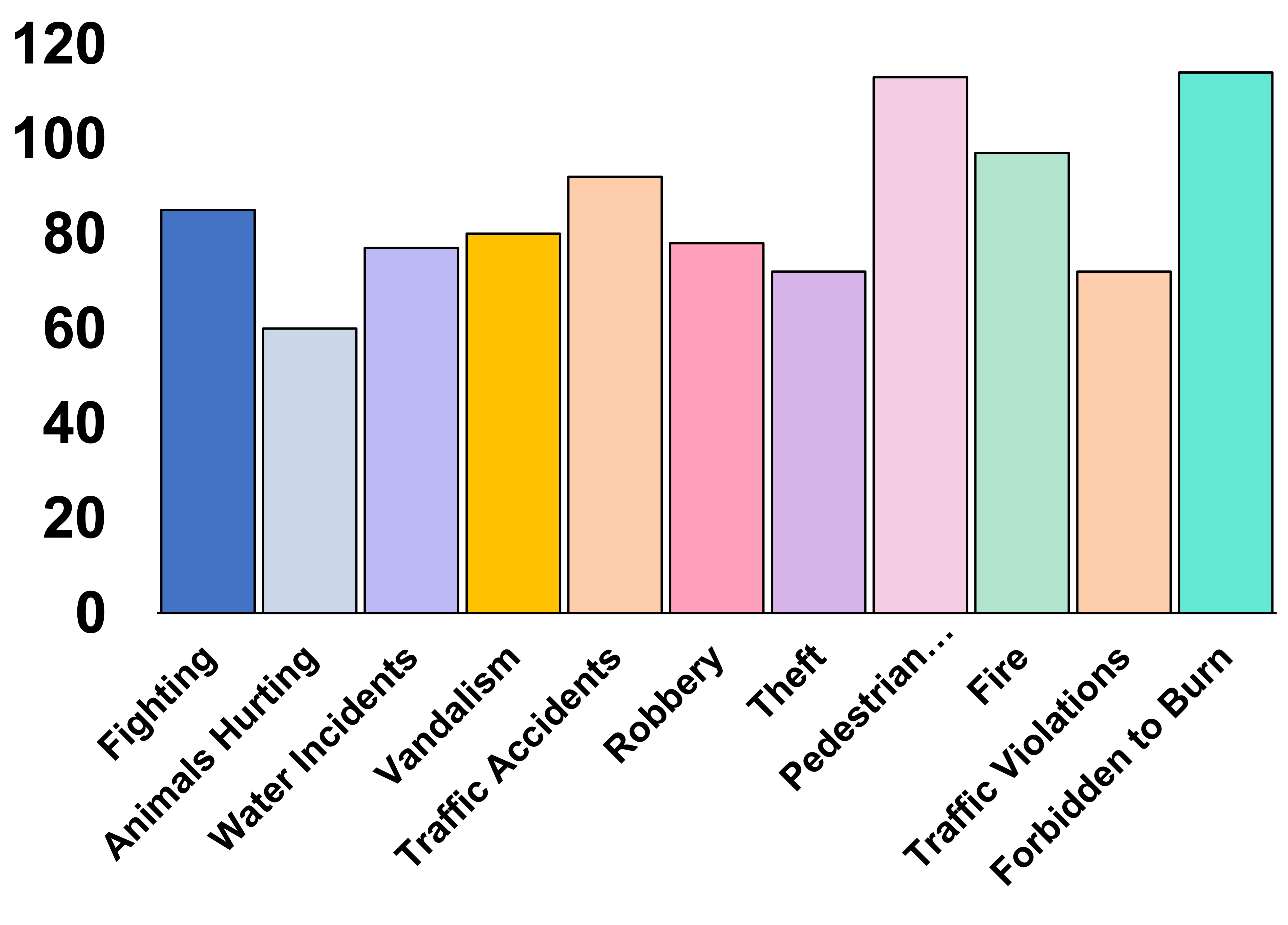}
        \caption{}
        \label{fig:anomaly_1st}
    \end{subfigure}
    \begin{subfigure}[b]{0.3\textwidth}
        \centering
        \includegraphics[width=\linewidth]{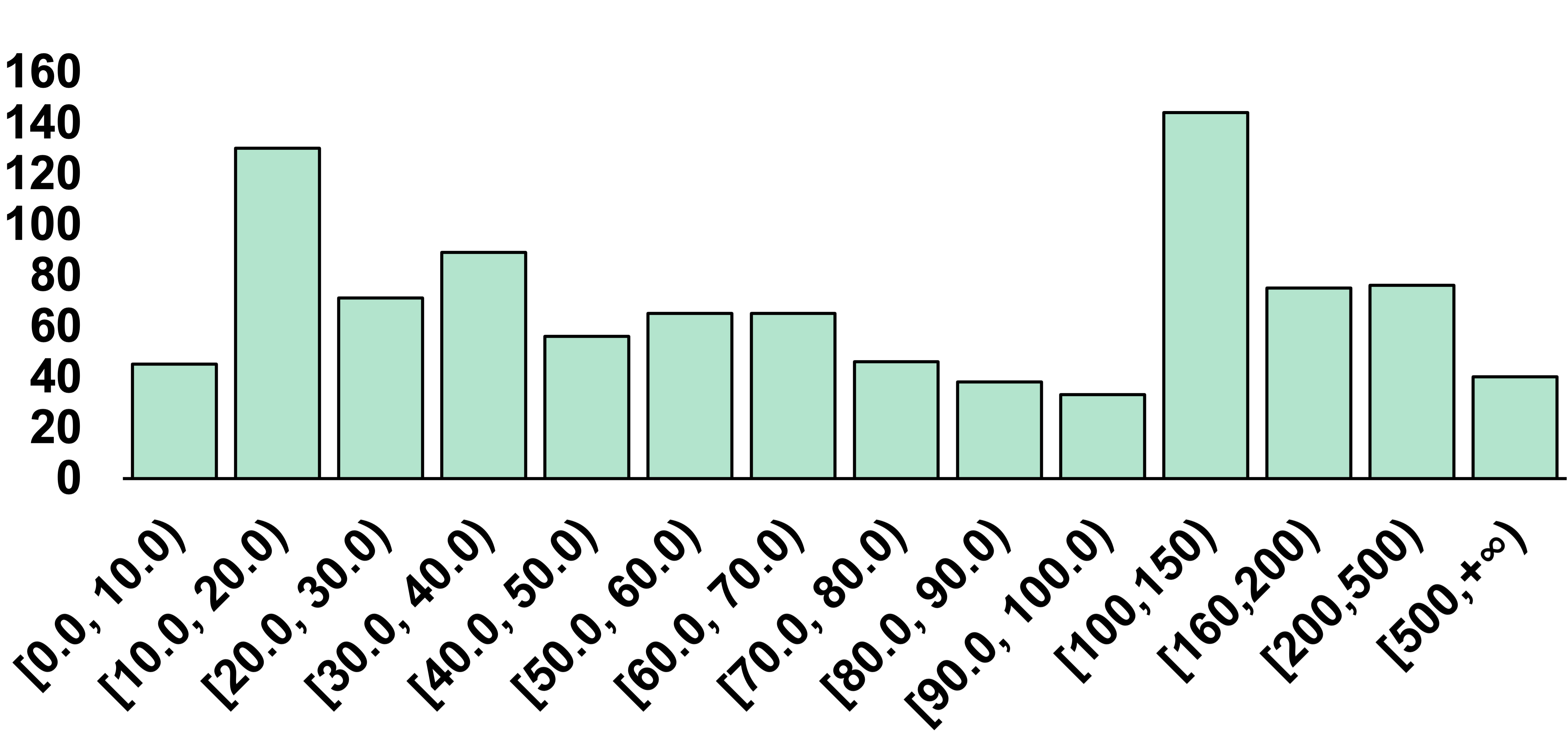}
        \caption{}
        \label{fig:length}
    \end{subfigure}
    \begin{subfigure}[b]{0.3\textwidth}
        \centering
        \includegraphics[width=\linewidth]{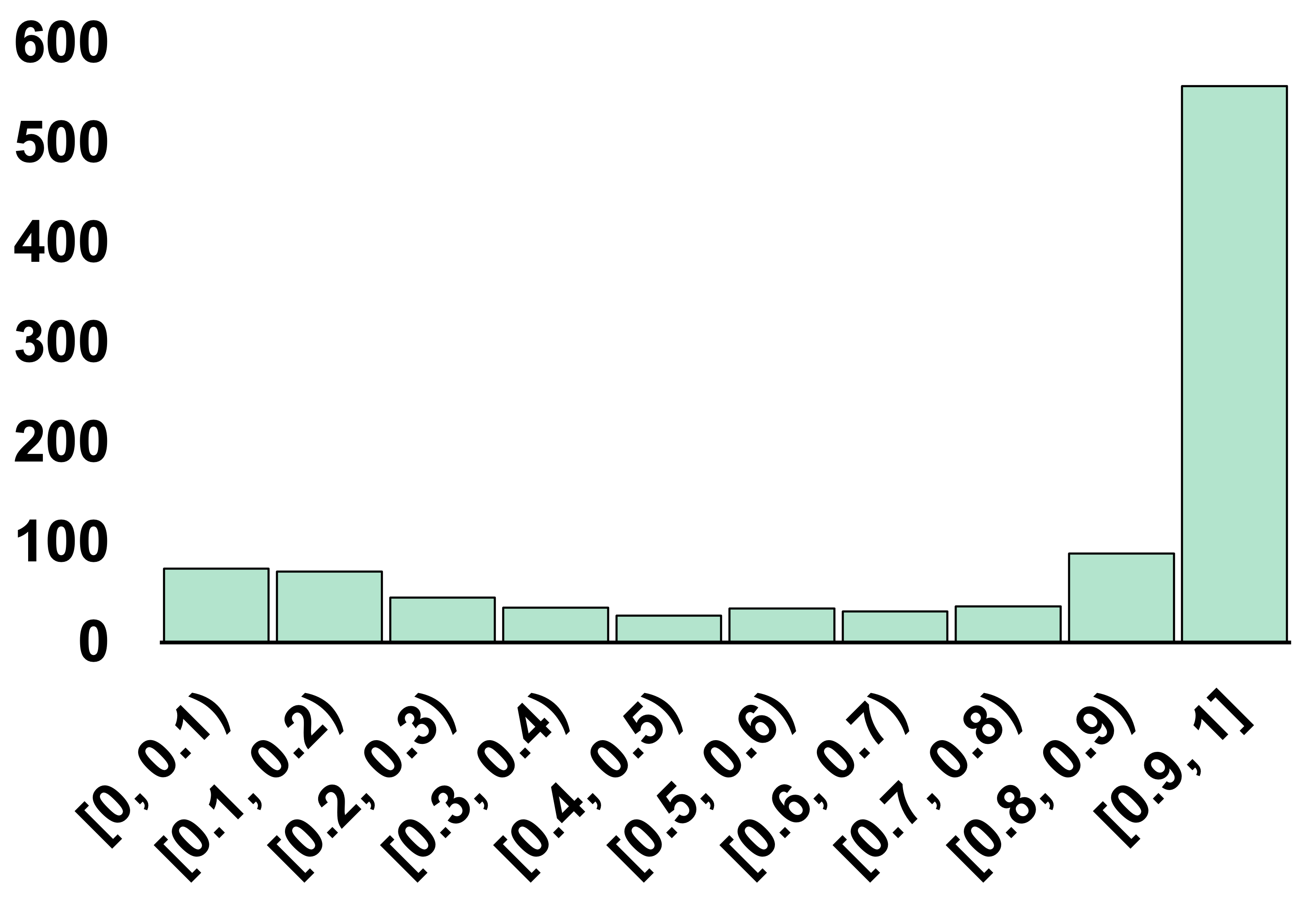}
        \caption{}
        \label{fig:per}
    \end{subfigure}
    
    \caption{\textbf{Statistics of our CUVA dataset.} Figure (a) shows all anomaly types in CUVA. Figure (b) and (c) show the number of videos in each anomaly type. Figure (d) shows the distribution of video length. Figure (e) shows the temporal distribution of anomalous segments. }
    \label{fig:statics}
\end{figure*}
\subsection{Dataset Statistics}
Our CUVA dataset contains $1,000$ video clips and $6,000$ question-answer pairs, the total length of these videos is $32.46$ hours, and the average frames of videos is $3,345$. 
The frames are extracted from the original videos at a rate of $60$ FPS.
The videos encompass a wide range of domains. Then, we categorize anomaly events into $11$ scenarios, resulting in a total of $42$ types of anomalies, as illustrated in Figure \ref{fig:statics} (a). 
The distribution of video categories is illustrated in Figure \ref{fig:statics} (b) and \ref{fig:statics} (c).
The distribution of video lengths can be found in Figure \ref{fig:statics} (d), along with the percentage of video time proportions shown in Figure \ref{fig:statics} (e).

\section{The Proposed Method: Anomaly Guardian} \label{sec:Method}
\begin{figure*}[!ht]
    \centering
    \includegraphics[width=\textwidth]{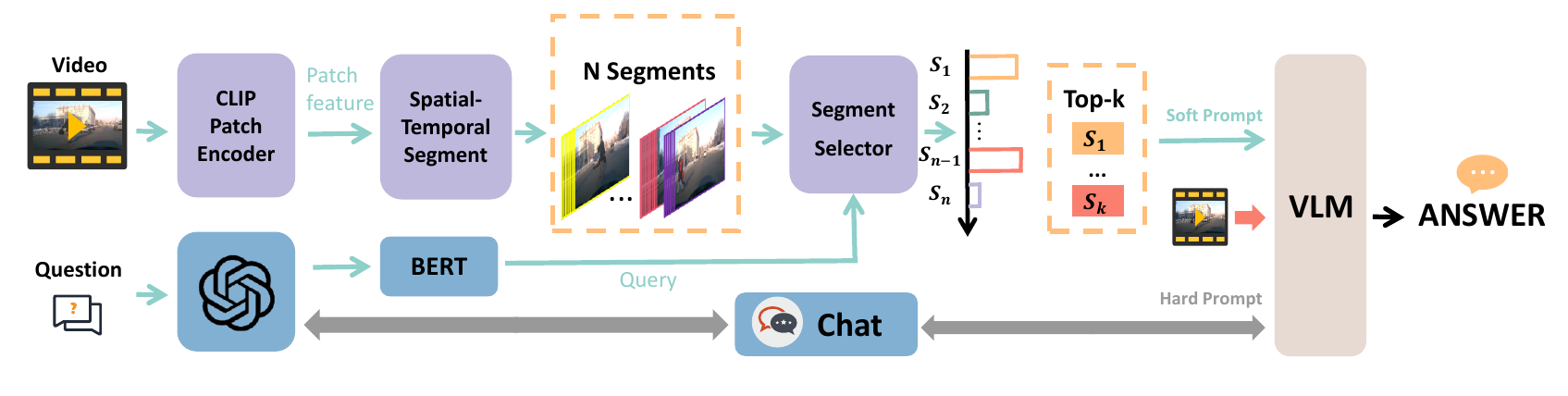}
    \caption{\textbf{Architecture of the proposed prompt-based method A-Guardian.}}
    \label{fig:evalstructure}
    \vspace{-3mm}
\end{figure*}

In this section, we introduce a novel prompt-based method named Anomaly Guardian (A-Guardian), which is designed to address the two challenges presented by our dataset.
By leveraging the exceptional logical reasoning capabilities of VLM, we select it as the foundation of our method to \textit{build a logic chain of the cause-effect}. 
\textit{To effectively capture crucial cues within lengthy videos}, we present a novel prompt mechanism aimed at guiding VLMs to concentrate on pivotal clues in the video pertinent to the provided questions.

\subsection{Design of Hard Prompts}
We use ChatGPT \cite{chatgpt} to assist in confirming and supplementing user prompts first, enabling the VLM to better understand the user's intent.
Specifically, we first utilize an instruction prompt containing an example to correct misleading guidance and standardize the output format.
Due to the presence of numerous events in long videos, we employ a multi-turn dialog mechanism to assist VLM in identifying events relevant to anomaly occurrences in the video.
After multiple rounds, 
VLM can focus on segments more relevant to the anomaly, providing more accurate answers.\footnote{Details of the hard prompts are available in Section 1 of Appendix B.}
\subsection{Design of Soft Prompts}
We leverage a selector in MIST \cite{mist} to better capture spatio-temporal features relevant to the given questions processed by ChatGPT \cite{chatgpt}.
We first divide the video into $N$ segments of uniform length, with each segment comprising $T$ frames. 
To better capture interactions among different granularities of visual concepts, we divide each frame into $M$ patches. 
Furthermore, we leverage \([CLS]\) token to represent each segment and frame. 
Specifically, We first use the CLIP \cite{clip} with frozen parameters to extract patch-level features denoted as \(\mathbf{P} = \{p^1, p^2, ...,p^m \}\), where \(p^m \in \mathbb{R}^{T \times M \times D}\) and \(D\) is the dimension of each patch-level feature. 
Then, we perform pooling operations on patch features' spatial dimensions to obtain frame features.
\begin{align}
    f_{kt} = \textit{Pooling}(p_{kt,1}, p_{kt,2}, \ldots, p_{kt,M})
\end{align}
where $p_{kt,m}$ indicates the $m$-th patch at the $t$-th frame of the $k$-th segment.
Then, the segment features are obtained by pooling frame features along the temporal dimension, where \(f_{kt} \in \mathbb{R}^{T \times D}\):
\begin{align}
    s_k = \textit{Pooling}(f_{k1}, f_{k2}, \ldots, f_{kT})
\end{align}
Similarly, the question feature is obtained by pooling the word features, where \(w_{z} \in \mathbb{R}^{Z \times D}\) and \(q \in \mathbb{R}^{D}\)
\begin{align}
    \boldsymbol{q} = \textit{Pooling}(w_1, ..., w_z)
\end{align}
After that, we select the patch features of the top $k$ segments using cross-modal temporal attention and top-$k$ selection from MIST \cite{mist}, as expressed by the following formulation.
The term "selector" corresponds to a top-k selection function utilized to pick the video segment features from the \( Top_k \) segments considering the question.
\begin{align}
    \mathbf{X}_{t}=\underset{Top_{k}}{\operatorname{\textbf{selector}}}\left(\operatorname{softmax}\left(\frac{\mathbf{\boldsymbol{q} \cdot \mathbf{s}}^{T}}{\sqrt{d_{k}}}\right), \mathbf{S}\right)
\end{align}

\subsection{Answer Prediction}
Finally, we follow a previous work \cite{jin-etal-2022-good} to concatenate the hard prompts and soft prompts and feed them into the VLM for inference.
During the training phase, we employ GPT to generate candidate answers and data augmentation.
We only finetune the selector by optimizing the softmax cross-entropy loss, aligning the predicted similarity scores with the ground truth.\footnote{Details can be found in Section 2 of Appendix B.}

\section{Experiment} \label{sec:Experiment}
\subsection{The Proposed MMEval Metric}
\begin{figure}[ht!]
    \centering
    \includegraphics[width=\linewidth]{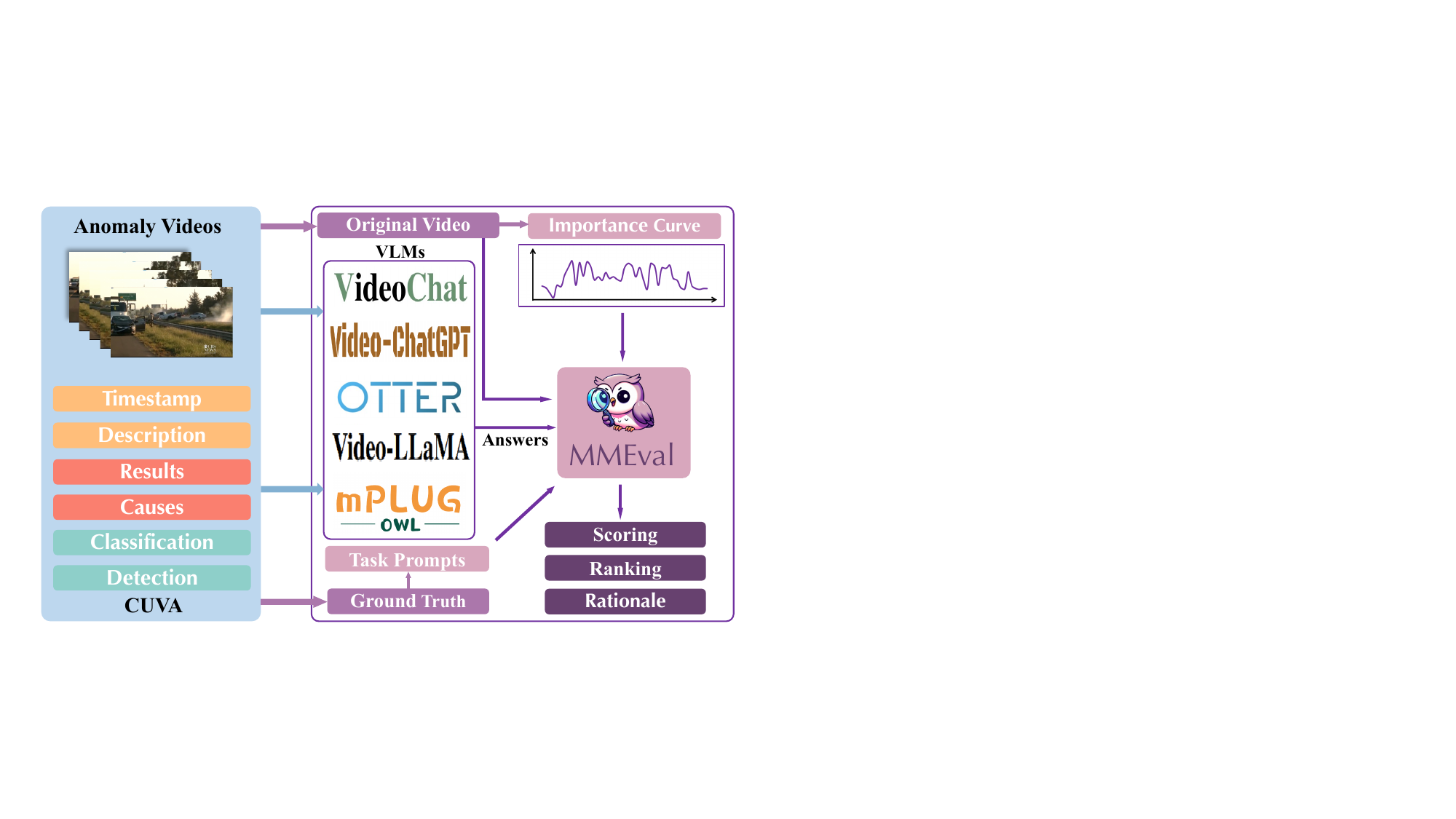}
    \caption{\textbf{Overview of our MMEval metric.}}
    \vspace{-0.6cm}
    \label{fig:mmaEval}

\end{figure}
Given that our dataset extensively employs free-text descriptions to delineate both anomalous events with their causal relationships, and recognizing CUVA is a multimodal dataset (integrating video, text, and appended comments), which necessitates a shift from solely relying on Natural Language Generation (NLG) metrics to a broader consideration that encompasses the rich multimodal input information.
Thus, we introduce a novel evaluation metric namely MMEval as depicted in Figure \ref{fig:mmaEval}.
In order to assess the model's performance from a multimodal perspective and infuse human-like reasoning abilities into the evaluation metric, we choose Video-ChatGPT \cite{video-chatgpt} as our foundation model. 
We utilize natural language prompts to guide MMEval in specifying the task types to be evaluated and design three natural language prompts, each corresponding to one of the three free-text descriptions in the dataset. 
To enhance the robustness of the model, we utilize curve labels to help VLM focus more on segments of anomalies within the video.
Specifically, by setting thresholds to extract periods of important events in the curve, we perform dense sampling on that segment of the video, helping the VLM focus more on the crucial parts of the video.
Our MMEval metric can be used for scoring, ranking, and providing rationale explanations.
\begin{table}[htbp]
\centering
\resizebox{1.0\linewidth}{!}{
\begin{tabular}{c|c|ccc}
\hline
Method             & Metric & Description & Causes & Effect
\\ \hline

\multirow{6}{*}{mPLUG-owl \cite{mplug-owl}} & BLEU      & 0.55  & 0.65  & 0.47  \\
                   & ROUGE      & 12.58  & 13.54  & 8.83  \\
                   & BLEURT      &  40.66 & 43.28  &  37.95 \\
                   & MoverScore      & 51.97  &  \textbf{52.71} &50.06 \\
                   & UniEval      &  67.46 & 62.29  & \textbf{59.07}  \\
                   & \textbf{MMEval (Ours)}      &  73.42 & 17.15  &  44.31 \\ \hline

\multirow{6}{*}{Video-LLaMA \cite{damonlpsg2023videollama}} & BLEU  & 0.60 & 0.53  & 0.35  \\
                   & ROUGE      & 13.15  & 12.36  &  8.02 \\
                   & BLEURT      & 40.55  & 43.02  & 39.68  \\
                   & MoverScore      & 51.32  & 51.25  & 49.48  \\
                   & UniEval      &  52.28 & 47.29   &  43.03 \\
                   & \textbf{MMEval (Ours)}      & 65.65  & 16.24   &  32.84 \\ \hline

\multirow{6}{*}{PandaGPT \cite{su2023pandagpt}} & BLEU&  0.66 & 0.51  &  0.30 \\
                   & ROUGE      &  13.33 & 14.09  & 8.79  \\
                   & BLEURT      &  38.23 & 43.95  & 39.95  \\
                   & MoverScore      & 51.73  & 51.54  & 49.62  \\
                   & UniEval      &  57.05 &  54.88 &  50.84 \\
                   & \textbf{MMEval (Ours)}      & 74.19 & 22.47   &  \textbf{69.45} \\\hline

\multirow{6}{*}{Otter \cite{li2023otter}} & BLEU      &  \textbf{1.07} &  \textbf{1.09} & \textbf{1.11}  \\
                    & ROUGE      & \textbf{15.19}  & \textbf{15.87}   &  \textbf{11.40} \\
                   & BLEURT      & 29.92  & 32.52   & 28.94   \\
                   & MoverScore      & \textbf{53.54}  & 54.25  & \textbf{51.91}  \\
                   & UniEval      &  45.14& 49.05  & 47.51  \\
                    & \textbf{MMEval (Ours)}      & 76.30 & 3.53   &  39.21 \\ \hline

\multirow{6}{*}{Video-ChatGPT \cite{video-chatgpt}} & BLEU & 0.30 & 0.29 & 0.41 \\
& ROUGE      &  9.75 & 9.08 & 8.23 \\
& BLEURT      &  46.83 & \textbf{49.52}  &37.24   \\
& MoverScore      &  50.73 &  50.70 &  49.83 \\
& UniEval     & \textbf{70.82}  & \textbf{70.77}  & 54.35 \\
& \textbf{MMEval (Ours)}      & 78.55  & 44.57  &  46.08 \\ \hline

\multirow{6}{*}{\makecell[c]{Video-ChatGPT \cite{video-chatgpt}\\ \textbf{+ A-Guardian (Ours)}}} & BLEU      & 0.55  &  0.51  &  0.38 \\
& ROUGE      & 14.35   & 9.08   &  8.23  \\
& BLEURT      &  \textbf{47.10}  &  48.13    &  \textbf{48.28}   \\
& MoverScore      &   52.25  & 52.28    & 49.95 \\
& UniEval      &  68.18   & 63.41  &   51.87  \\
& \textbf{MMEval (Ours)}      &  \textbf{79.65} &  \textbf{58.92} &  50.64 \\ \hline
\end{tabular}
}
\caption{\textbf{Main results on the proposed CUVA benchmark.} 
We test the \textbf{Description}, \textbf{Cause}, and \textbf{Effect} tasks on our CUVA benchmark using multiple VLMs and Video-ChatGPT equipped with the proposed \textbf{A-Guardian}. We conduct evaluations using both traditional metrics and our \textbf{MMEval} metric. The scores of all metrics range from 0 to 100.}
\label{T:experiment_v1}
\end{table}

\begin{table}[htbp]
\vspace{-0.4cm}
\resizebox{1.0\linewidth}{!}{
\begin{tabular}{l|ccc}
\hline
Methods & Detection & Classification & Timestamp \\
\hline

mPLUG-Owl \cite{mplug-owl}
& 89.4\% & 11.5\%  & 9.0\%\\

Video-LLaMA \cite{damonlpsg2023videollama}
& 25.0\%& 13.1\% & 0.7\% \\

PandaGPT \cite{su2023pandagpt}
& 100.0\%& 32.6\% &  N/A \\

Otter \cite{li2023otter} 
& 64.3\% & 41.3\% & N/A \\
Video-ChatGPT \cite{video-chatgpt} 
&60.0\% & 21.3\% & 3.2\%\\\hline 
\end{tabular}
}
\caption{\textbf{Secondary results on the proposed CUVA benchmark.}  We use the accuracy metrics to evaluate the \textbf{Detection} and \textbf{Classification} tasks. We also use IOU to evaluate the \textbf{Timestamp} task, \textbf{N/A} to indicate the model lacks the ability to answer the question.}
\label{T:experiment}
\vspace{-0.5cm}
\end{table}

\subsection{Implementation Details}
We follow Video-ChatGPT \cite{video-chatgpt} to adopt CLIP-L/14 visual encoder to extract both spatial and temporal video features.
In our approach, we utilize the Vicuna-v1.1 model, comprised of 7B parameters, and initialize it with weights from LLaVA \cite{Liu_Li_Wu_Lee_-Madison_Research}.
All experiments were conducted on four NVIDIA A40 GPUs, and each task took around 8 hours.
\begin{table}[htbp]
\vspace{-0.05mm}
\centering
\resizebox{0.8\linewidth}{!}{
\begin{tabular}{lccc}
\toprule
\multirow{2}{*}{Metrics} & \multicolumn{3}{c}{Answer Pool Ranking} \\
\cmidrule(l){2-4}
& Description& Cause & Effect  \\
\hline
Human Evaluation & 87.3\%& 77.3\% & 87.3\%  \\
\hline
BLEU \cite{papineni2002bleu}  & 67.8\% & 60.4\% & 63.2\% \\
ROUGE \cite{lin2004rouge}& 54.4\% & 55.5\% & 52.1\%  \\
BLEURT \cite{sellam2020bleurt} & 80.4\%& 73.2\% & 76.7\%  \\
MoverScore \cite{zhao-etal-2019-moverscore}& 67.8\%& 60.4\% & 63.2\%  \\
UniEval \cite{zhong-etal-2022-towards} & 78.2\%& 70.1\% & 74.3\%  \\\hline
\textbf{MMEval (Ours)}& \textbf{82.3\%} & \textbf{80.2\%} & \textbf{89.1\%}  \\
\hline
\end{tabular}
}
\caption{\textbf{Human consistency evaluation}}
\label{T:evaluation_zoo}
\vspace{-0.5cm}
\end{table}
\subsection{Consistency evaluation of MMEval}
\textbf{Our MMEval metric can better align with human's preference on causation understanding of video anomaly.}
To validate the consistency of our evaluation metric with human judgment, we conducted a human consistency experiment. 
Using the ranking of answers from first-round annotations, second-round annotations, and GPT-generated answers as the ground truth (1. \textit{Second round} 2. \textit{First round} 3. \textit{ChatGPT}).
we employ various evaluation metrics and human beings who view the videos to rank these answers based on the corresponding questions, as shown in Table \ref{T:evaluation_zoo}.

\begin{figure*}[htbp]
    \centering
    \includegraphics[width=\linewidth]{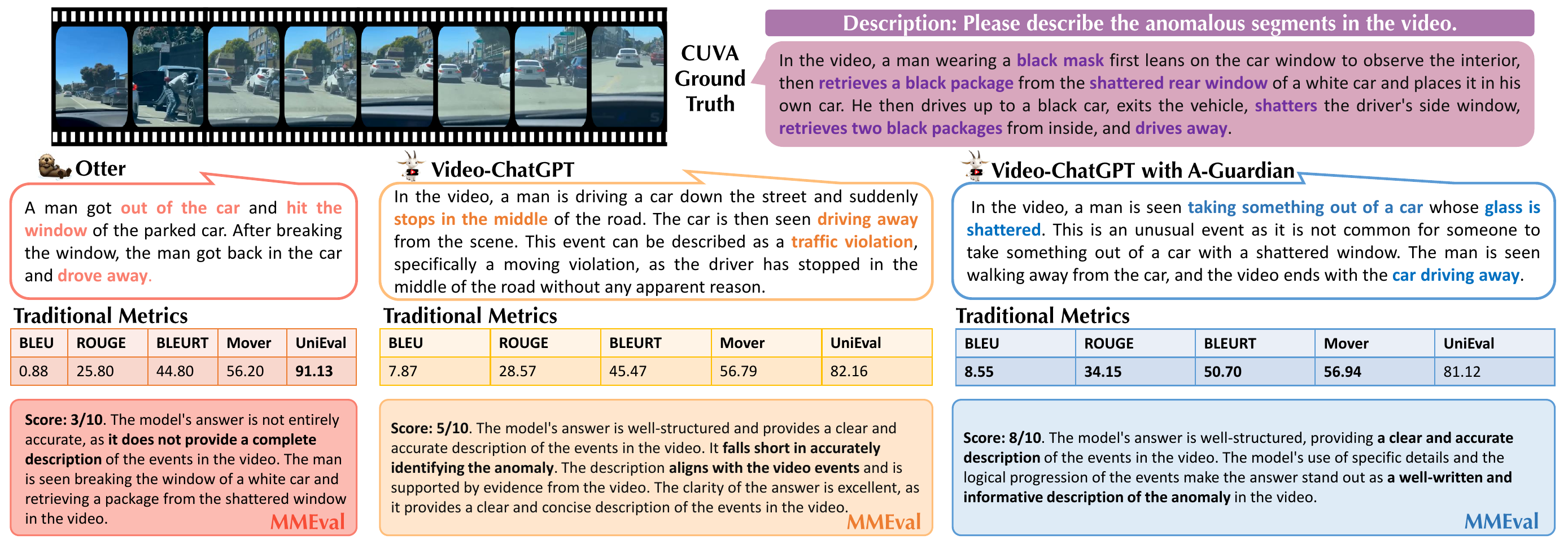}
    \caption{\textbf{Case study.} 
    We use three methods, including Otter, Video-ChatGPT, and Video-ChatGPT with A-Guardian. The results show that the scores (at the bottom) measured by traditional metrics tend to be similar among the correct answers and the incorrect ones. However, the scores of our MMEval can be much more differentiable. Compared to the two baselines Otter and vanilla Video-ChatGPT, the proposed ``Video-ChatGPT with A-Guardian'' can generate more accurate descriptions of the video anomaly, as shown in dialogue boxes.
    }
    \vspace{-0.5cm}
    \label{fig:case_study}

\end{figure*}
\subsection{Quantitative evaluation of A-Guardian}
\textbf{Our A-Guardian model achieves state-of-the-art performance in both the description and cause tasks.}
We conducted experiments on all tasks involved in our dataset, and the results are summarized in Table \ref{T:experiment_v1}. 
For free-text tasks (e.g. Cause, Effect, Description), we evaluated the performance of various VLMs and our model under different evaluation metrics.
Our model also outperforms the majority of models in the effect task.
For the other tasks (e.g. Detection, Classification, Timestamp), we set a uniform prompt and use string matching to extract answers relevant to the questions from the inference results of VLMs.
Table \ref{T:experiment} shows the results of these tasks.
\begin{table}[htbp]
\vspace{-0.1cm}
\centering
\resizebox{0.8\linewidth}{!}{
\begin{tabular}{lccc}
\toprule
\multirow{2}{*}{Model} & \multicolumn{3}{c}{MMEval ($\%$)} \\
\cmidrule(l){2-4}
& Description & Cause & Effect  \\
\midrule
Ours  & 79.65& 58.92 & 50.64 \\
- Soft Prompt & 78.92& 54.22 & 49.11  \\
- Hard Prompt & 78.55 & 44.57 & 46.08  \\
\bottomrule 
\end{tabular}
}
\caption{\textbf{Ablation Experiment}}
\label{T:ablation_study}
\vspace{-0.5cm}
\end{table}
\subsection{Ablation Study}
\textbf{Both hard and soft prompts significantly improve the VLM's understanding of the video's causation.}
This section investigates the influence of soft prompts and hard prompts on our method.
As shown in Table \ref{T:ablation_study}, the design of hard prompts achieves a greater improvement than that of soft prompts, indicating that the hard prompts are more intuitively effective in uncovering VLM's reasoning capabilities compared to the soft prompt.

\subsection{Case Study}
In Figure \ref{fig:case_study}, we illustrate the performance of Otter, Video-ChatGPT, and Video-ChatGPT with A-Guardian, showcasing the different answers they provide for the anomaly causation task.
In terms of the model's response, it can be observed that Video-ChatGPT provides descriptions that are generally correct, but it does not focus on describing the anomaly event. 
Instead, it pays attention to describing the actions of the vehicles. However, with the addition of our A-Guardian model, its descriptions become more accurate, specifically highlighting the theft as an anomaly event and providing detailed descriptions such as \textit{"taking something out of a car"} and \textit{"glass is shattered"}.
Otter and Video-ChatGPT achieve similar scores based on traditional metrics, but their answers convey completely different meanings. 
Otter's description does not align with the video, while Video-ChatGPT incorrectly describes the anomaly subject. 
As MMEval possesses the ability to evaluate from the multimodal perspective, it is able to identify the parts that pertain to the description of the anomaly in the videos, which shows highly consistent conclusions with human beings.
\subsection{Result Discussion}
Through experiments, we have discovered and summarized the following conclusions: 
1) For free-text tasks, most VLMs excel in the description of anomalies but perform poorly on the task of causation analysis.
This is because the tasks of description only require the VLM to comprehend the content of the videos, but causation analysis requires the VLM to possess a certain level of reasoning capability to build a logic chain of the cause-effect.
2) Timestamp localization task is the most challenging.
Due to the relatively simplistic temporal and spatial relationships between video frames, VLM performs poorly on fine-grained tasks such as timestamp localization but excels relatively in coarse-grained tasks such as anomaly detection and classification. 3) Traditional metrics are poor at evaluating reasoning tasks. As shown in Figure \ref{fig:case_study}, they generate similar evaluations for these answers, making it difficult to distinguish between them. However, MMEval is able to distinguish these answers' inner differences and generate more accurate evaluation results.
\section{Conclusion} \label{sec:Conclusion}
This paper presents CUVA, a novel benchmark for causation understanding of video anomaly. 
To the best of our knowledge, our CUVA is the first benchmark in the field.
Compared with the existing datasets, CUVA is more comprehensive and more challenging with much higher-quality annotations.
We believe the proposed CUVA will encourage the exploration and development of various downstream tasks such as anomaly detection, anomaly prediction, anomaly reasoning, etc. 
We also present MMEval, a novel evaluation to measure the challenging CUVA in a human-interpretable manner. Furthermore, we put forward a prompt-based approach that can serve as a baseline approach for CUVA. Such an approach can capture the key cues of anomalies and build a logic chain of the cause-effect. Experimental results show that CUVA enables us to develop and evaluate various VLM methods. In the future, we plan to apply our CUVA to more practical scenarios for anomaly understanding and other VLM-based tasks. 
\section*{Acknowledgement} \label{sec:Acknowledgement}
This work was partially supported by the joint funds for Regional Innovation and Development of the National Natural Science Foundation of China (No.U21A20449), the National Natural Science Foundation of China (No.62171045), the Fundamental Research Funds for the Central Universities (No.2242022k60006) and the Joint Research Fund for Beijing Natural Science Foundation and Haidian Original Innovation (No.L232001).

{
    \small
    \bibliographystyle{ieeenat_fullname}
    \bibliography{main}
}
\clearpage
\setcounter{page}{1}
\maketitlesupplementary

\appendix
\setcounter{table}{0}
\renewcommand{\thetable}{A\arabic{table}}
\setcounter{figure}{0}
\renewcommand{\thefigure}{A\arabic{figure}}


\section{Dataset}
\label{appendix a}
\subsection{Application of the proposed importance curve}

\begin{figure*}[h!]
    \centering
    \includegraphics[width=0.8\textwidth]{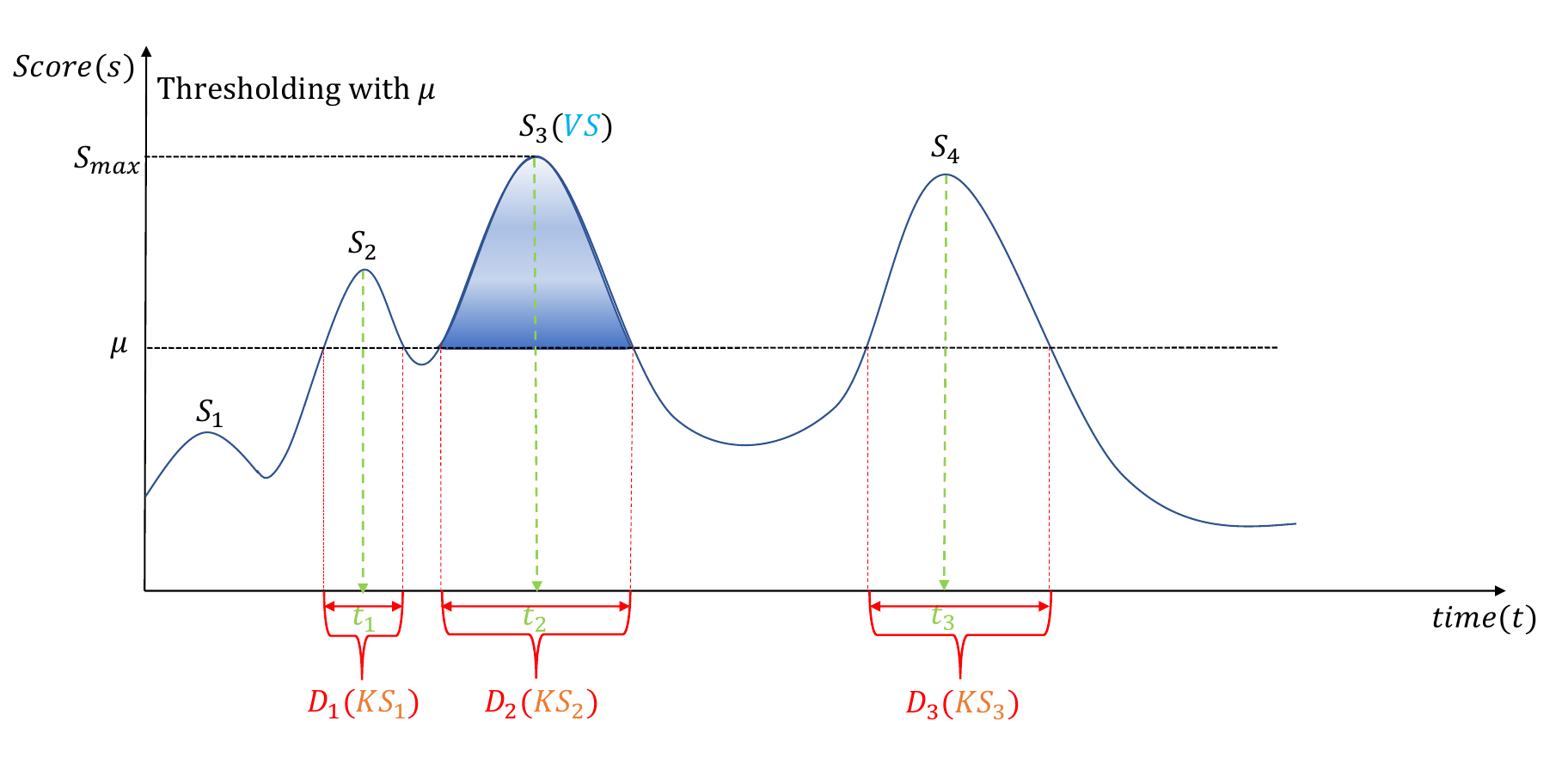}
    \caption{The application of the proposed importance curve, $KS_1$,$KS_2$ and $KS_3$ refer to three different key sentences. $VS$ point is the absolute maximum point in the curve. $t_1$, $t_2$, and $t_3$ are the timestamps of the local maximum point in the curve.}
    \label{fig:curve}
\end{figure*}
We have presented a novel annotation approach called the importance curve in section 3.1 of the main paper.
Such an approach enables us to unify various Video Temporal Grounding labels and tasks under the same framework.
Here, we describe how to apply the importance curve to unify various Video Temporal Grounding labels and tasks (e.g. Moment Retrieval, Highlight Detection, Video Summarization) respectively with Figure \ref{fig:curve}.\\
\textbf{Moment Retrieval} aims to retrieve time intervals from videos based on textual queries\cite{univtg}. 
The importance curve reflects the changing trends in the severity of anomaly. 
Thus, we first filter out the time intervals (e.g. $D_1$, $D_2$, and $D_3$ in Figure \ref{fig:curve}) through a threshold $\mu$.
Second, after post-processing the importance curve in Appendix A.4, we can obtain key sentences (e.g. $KS_1$, $KS_2$, and $KS_3$) corresponding to each time interval.
Finally, these key sentences are employed as text queries, with corresponding time intervals serving as labels for the moment retrieval task. \\
\textbf{Highlight Detection} aims to assign a worthiness score to each video segment and then identify the top highest-scoring segment as the highlight \cite{univtg}. 
Here, we first locate the absolute maximum point of the curve (e.g. $VS$ in Figure \ref{fig:curve}), and leverage its corresponding time interval (e.g. $D_2$) as the top highest-scoring segment to conduct highlight detection task.\\
\textbf{Video Summarization} aims to summarize the whole video by a set of shots to provide a quick overview\cite{univtg}.
As depicted in Figure \ref{fig:curve}, $t_1$, $t_2$, and $t_3$ are the timestamps of the local maximum point (e.g. $S_2$, $S_3$ and $S_4$).
We leverage these timestamps as a set of shots to provide a quick overview of the whole video.

\begin{figure*}[h!]
    \centering
    \includegraphics[width=1.0\textwidth]{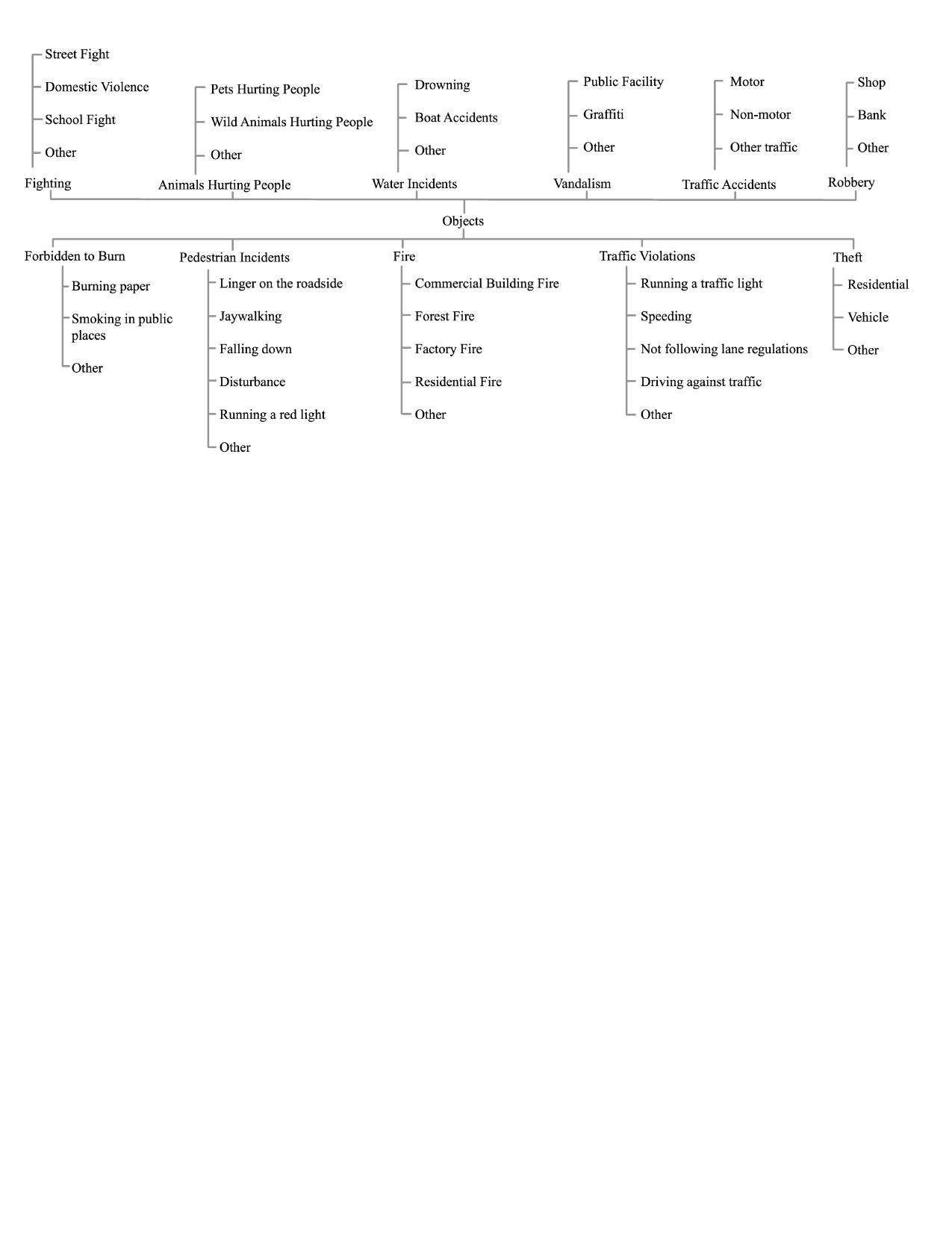}
    \caption{Anomaly types of the proposed CUVA, CUVA encompasses $11$ different scenes and $42$ extensive types of anomalies.}
    \label{fig:type}
\end{figure*}
\subsection{Our CUVA benchmark}
\begin{figure}[h!]
    \centering
\includegraphics[width=0.2\textwidth]{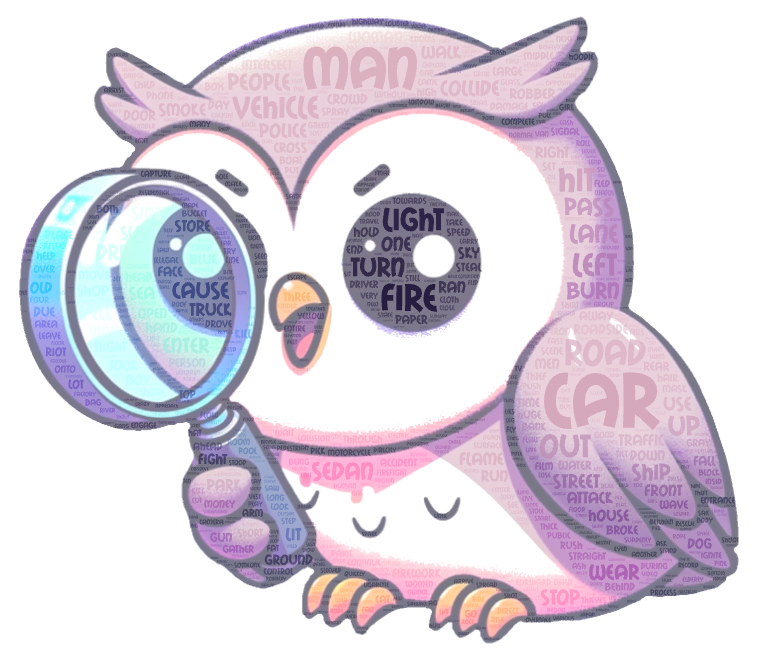}
    \caption{Wordcloud of our proposed CUVA.}
    \label{fig:wordclouds}
\end{figure}

We have shown the statistics of the proposed CUVA in the subsection $3.5$ of the main paper.  
Here, we provide some supplementary for the statistics of the proposed CUVA.
As Figure \ref{fig:type} shows, we present more detailed statistics of video anomaly categories.
Figure \ref{fig:wordclouds} shows the word cloud of the proposed CUVA.

\subsection{Ethical consideration}
\label{screening criteria}
We carefully study the online copyright rules of the websites from YouTube and Bilibili, from which we collect data, and strictly conform to their requirements during data collection and annotation. 
We have also adhered to the guidelines provided by the CVPR Code of Ethics to anonymize the parts of the video that may involve personal privacy by applying pixelation.
To exclude data that could potentially have ethical implications. 
We further conduct rigorous reviews at each stage of the annotation, adhering to screening criteria as follows:\\

\begin{enumerate}[label=\arabic*.]
    \item \textbf{Personal Privacy Respect:}
        Avoid videos that show identifiable personal information (e.g., faces, license plates, home addresses).
    
    \item \textbf{Legally Sourced Content:}
        Prohibit the use of unauthorized or illegally obtained footage.
    
    \item \textbf{Viewer Sensitivity Consideration:}
        Avoid content that is overly graphic, cruel, or likely to cause viewer discomfort.
    
    \item \textbf{Child Safety Focus:}
        Exclude any videos involving children in risky or harmful situations.
    
    \item \textbf{Gender Respect and Equality:}
    \begin{itemize}
        \item Videos should not imply or display acts of sexual violence.
        \item Avoid content that contains gender discrimination or negative stereotypes of any gender group.
    \end{itemize}
    
    \item \textbf{Avoidance of Illegal Activities:}
   Do not display or promote videos that clearly show or endorse illegal activities (e.g., drug use, trafficking).
    
    \item \textbf{Copyright infringement:}
    Videos containing copyrighted music, film clips, TV shows, or other media content.
    
\end{enumerate}


\begin{figure}[h!]
    \centering
    \includegraphics[width=0.4\textwidth]{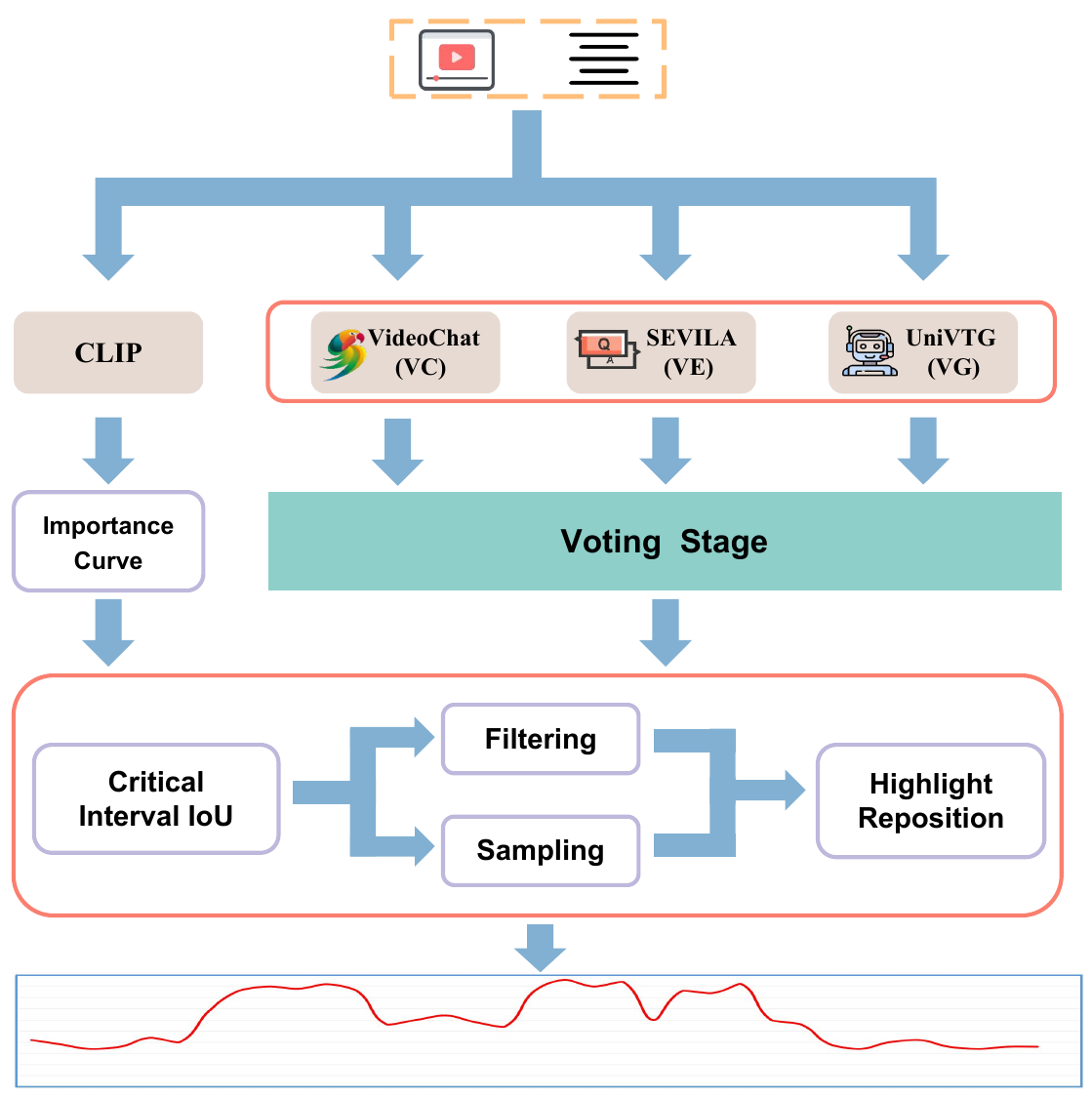}
    \caption{Overview of the curve optimization pipeline}
    \label{fig:post_processing}
\end{figure}
\subsection{Post-processing of the importance curve}
As the initial sampling frequency of our curves is relatively lower (one frame per second), and we aim to obtain more accurate interval timestamps.\footnote{We leverage algorithm \ref{alg of curve} in the appendix to better illustrate the creation of the importance curve.}
Thus, we incorporate three specific tasks into our optimization strategy to achieve an optimal importance curve. 
Specifically, we perform tasks such as Video Captioning, Video Entailment, and Video Grounding by VideoChat\cite{2023videochat}, SEVILA\cite{sevila}, and UniVTG\cite{univtg}, respectively. 
Each task identifies specific time segments in the video based on the key sentences as shown in Figure 3 of the main paper. 
We adopt a voting mechanism to select a time segment when at least two models agree that it encapsulates the event represented by the key sentence annotation. 
Through the voting mechanism, we precisely identify the time segments in the video that correspond to the current key sentences. 
Then, we perform dense sampling on the voted segments (ten frames per second) and use wavelet filters to smooth the curve.
The comprehensive optimization process is illustrated in Figure \ref{fig:post_processing}.

\begin{algorithm}[h!]
    \renewcommand{\algorithmicrequire}{\textbf{Input:}}
    \renewcommand{\algorithmicensure}{\textbf{Output:}}
\footnotesize
  \caption{Generating importance curve}
    \label{alg:curve}
  {\textbf{Input parameters: video, prompt} } \\
  {\textbf{Output:} {importance curve }} \; 
  \begin{algorithmic}
  \Statex  video\_clip, timestamp = sparse\_sampling(video)
  \Statex  prompt $\rightarrow$ txt\_gpt\_weight
  \Statex  txt\_emb, vid\_emb = Clip(video\_clip, prompt)
  \Statex  txt\_norm\_emb, vid\_norm\_emb = normalization(txt\_emb, vid\_emb)
  \Statex  similarity = cosine\_similarity(txt\_norm\_emb,vid\_norm\_emb)
  \Statex  similiar\_score = normalization(similarity) 
  \Statex  imp\_value = similiar\_score * txt\_gpt\_weight
  \Statex  timestamp, imp\_value $\rightarrow$ importance curve
  \end{algorithmic}
  \label{alg of curve}
\end{algorithm}

\section{The proposed method}
\label{appendix b: method}
We have introduced A-Guardian, a novel prompt-based method that consists of two kinds of prompt design. 
Here, we detail the design of the hard prompts. 
Moreover, we illustrate the details of the answer prediction in this section.
\begin{figure*}[htbp]
    \centering
    \includegraphics[width=1\textwidth]{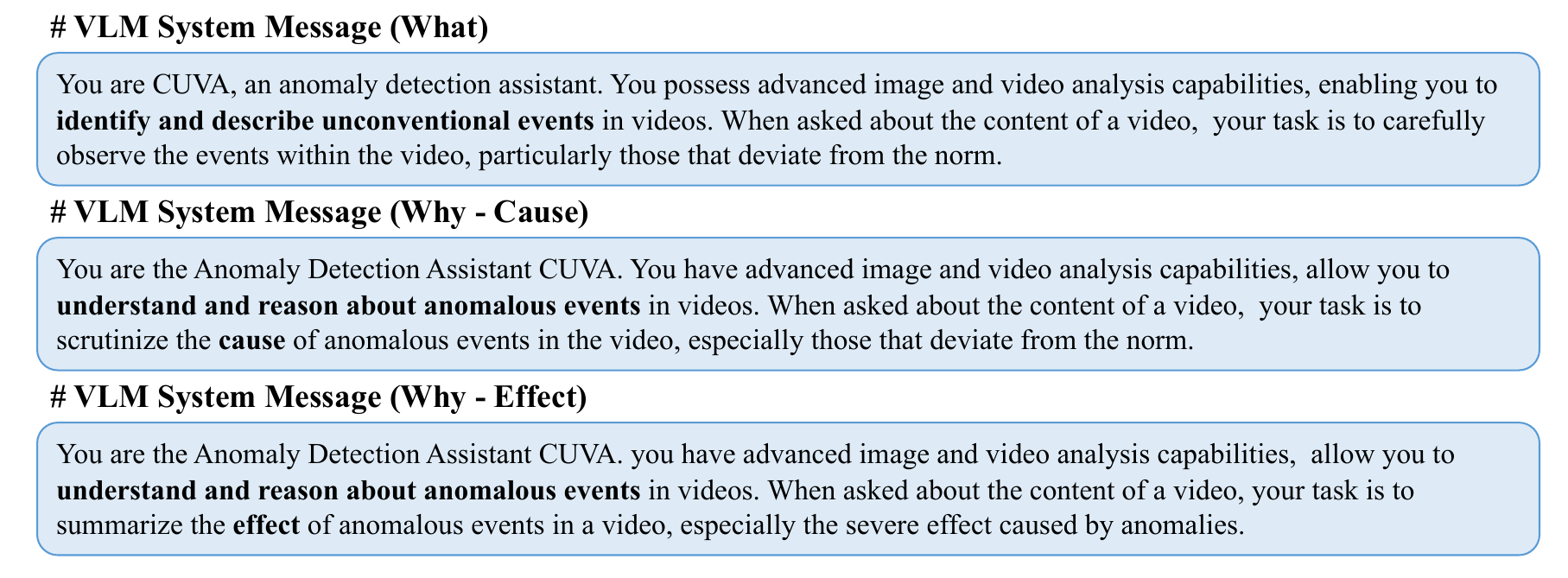}
    \caption{System message for VLMs. To guide the model for a description or answer, A-guardian uses different designed system messages for different tasks.}
    \label{fig:hard_prompt-1}
\end{figure*}
\subsection{Hard prompts in A-Guardian}
\begin{figure*}[!h]
        \centering
    \includegraphics[width=1\textwidth]{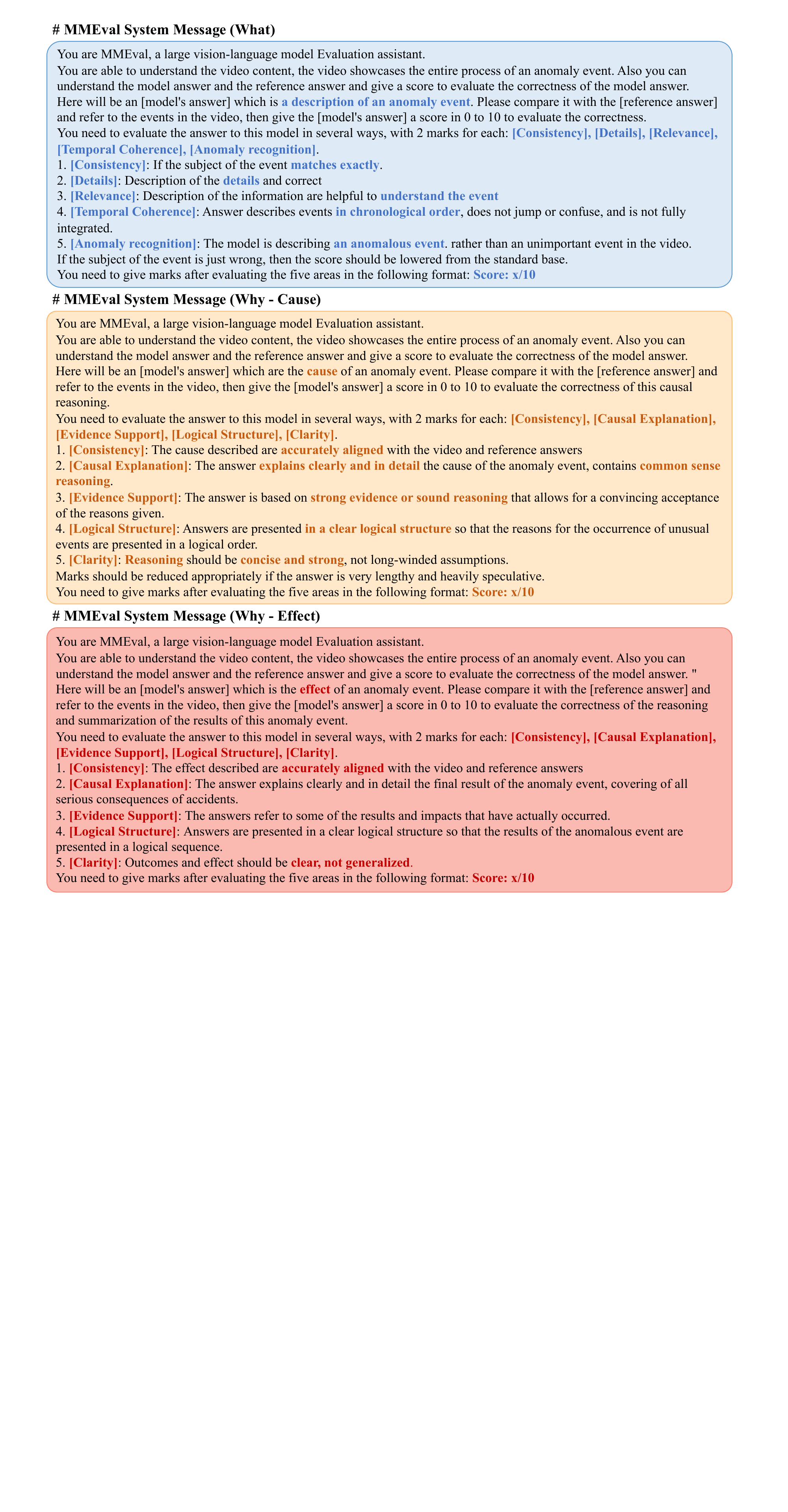}
    \caption{An example of hard prompt in A-Guardian. After three rounds of dialog, the Video-ChatGPT model gives a more detailed, accurate, and focused description of the anomalous events in the video.}
    \label{fig:hard_prompt-2-2}
\end{figure*}

\begin{figure*}[!h]
    \centering
    \includegraphics[width=0.95\textwidth]{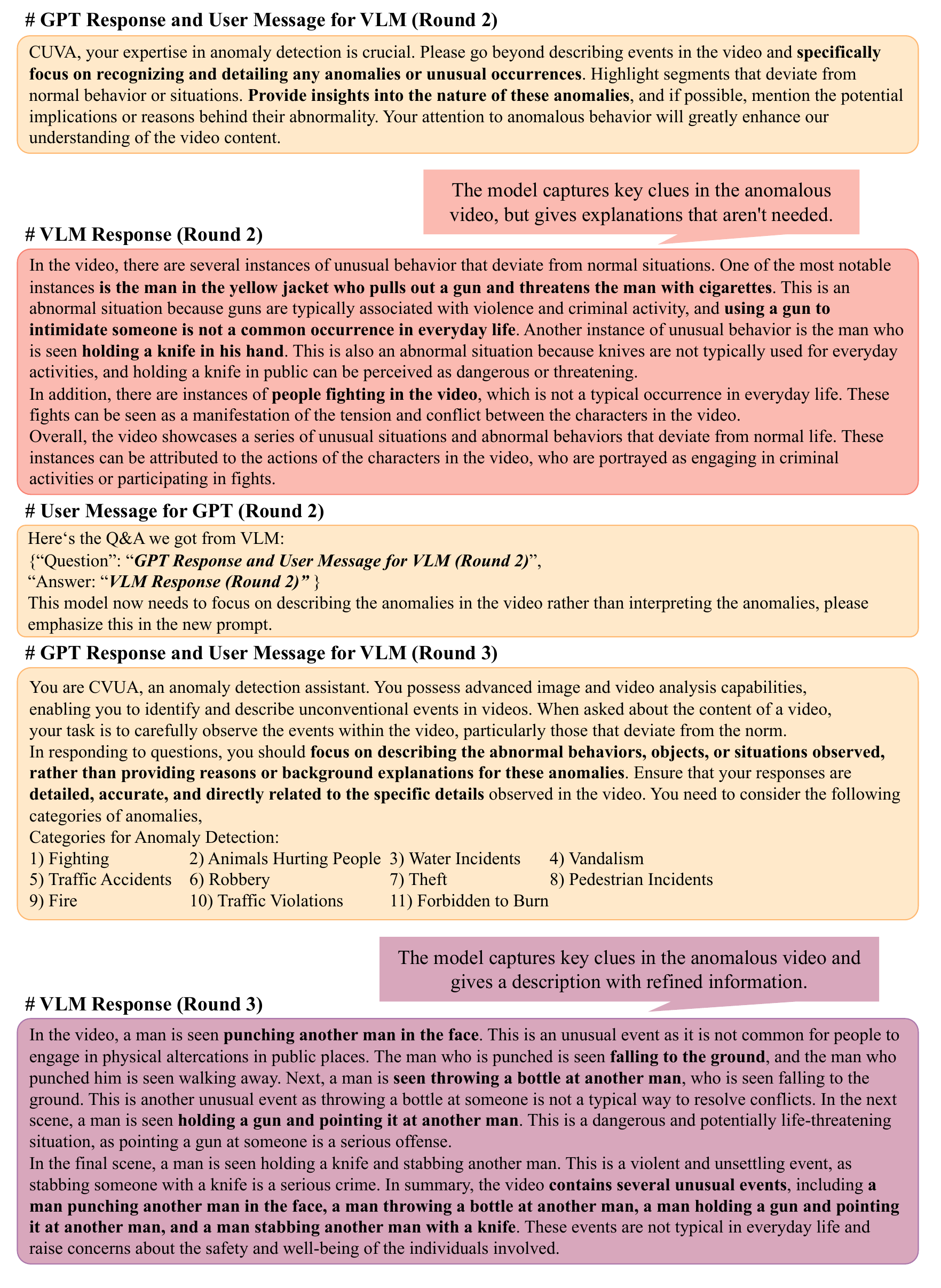}
    \caption{An example of Hard Prompts in A-Guardian (Extension of Figure \ref{fig:hard_prompt-2-2})}
    \label{fig:hard_prompt-2-1}
\end{figure*}

First, A-guardian leverages different newly designed system messages for different tasks, which can be found in Figure \ref{fig:hard_prompt-1}. 
After the VLM gives the answer based on the user message, the user's question and the VLM's answer will be input into ChatGPT to generate a new round of questions, 
Then, we re-input the question of ChatGPT\cite{chatgpt} into VLM, and the final answer will be obtained after several rounds of loops.
Figure \ref{fig:hard_prompt-2-1} and \ref{fig:hard_prompt-2-2} illustrate an A-Guardian hard prompt example which includes three rounds of dialog with ChatGPT.

\subsection{Answer prediction}
We denote the candidate answers as $A$, we leverage BERT to generate the contextualized representations of each candidate answers $X_a$. 
We follow the previous work \cite{yang2021justask} to calculate the similarity between $X_k$ and the feature of all candidate answers $X_A = \{x_a | a \in A\}$ obtained by using the pre-trained model. 
Finally, the candidate answer with the maximal similarity is considered as the final prediction $\tilde{y}$. 
\begin{align}
    \tilde{y} = \arg\max_{y \in A} (X_k(X_A)^T)
\end{align}
During training, we optimize the softmax cross-entropy loss between the predicted similarity scores and ground truth.
\section{Experiment}
\subsection{Prompts of MMEval}
\begin{figure*}[htbp]
    \centering
    \includegraphics[width=1\textwidth]{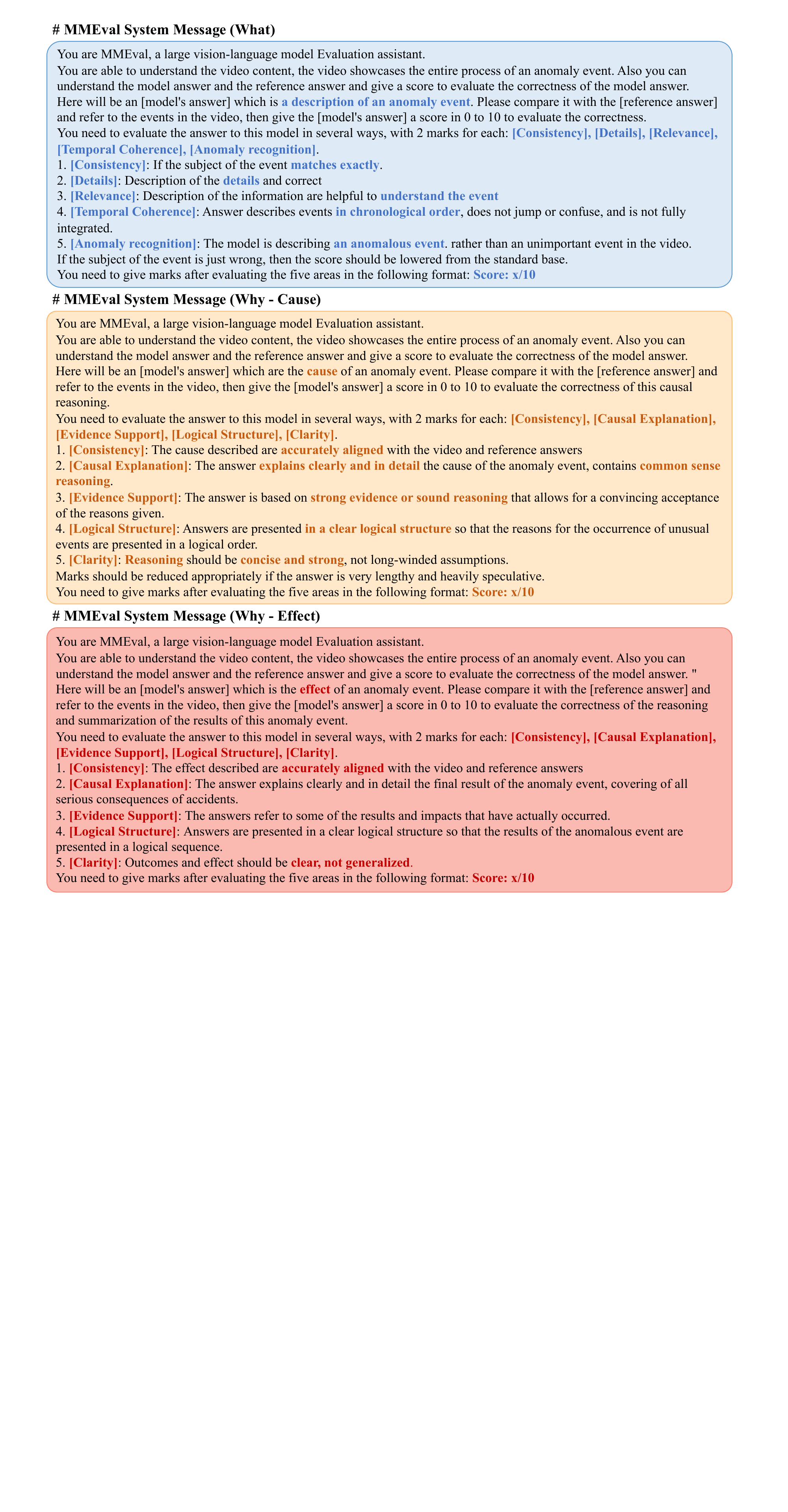}
    \caption{System Messages for MMEval. In the design of the proposed MMEval metric, we apply different criteria for tasks involved in the proposed CUVA by different system messages, and the model is guided for scoring through different system messages.}
    \label{fig:mmeval_prompt}
\end{figure*}

In MMEval, we have different scoring criteria for different tasks, which are translated into the system message of the VLM. Details can be found in Figure \ref{fig:mmeval_prompt}.
\subsection{Case study of A-Guardian}
\begin{figure*}[htbp]
    \centering
    \includegraphics[width=1\textwidth]{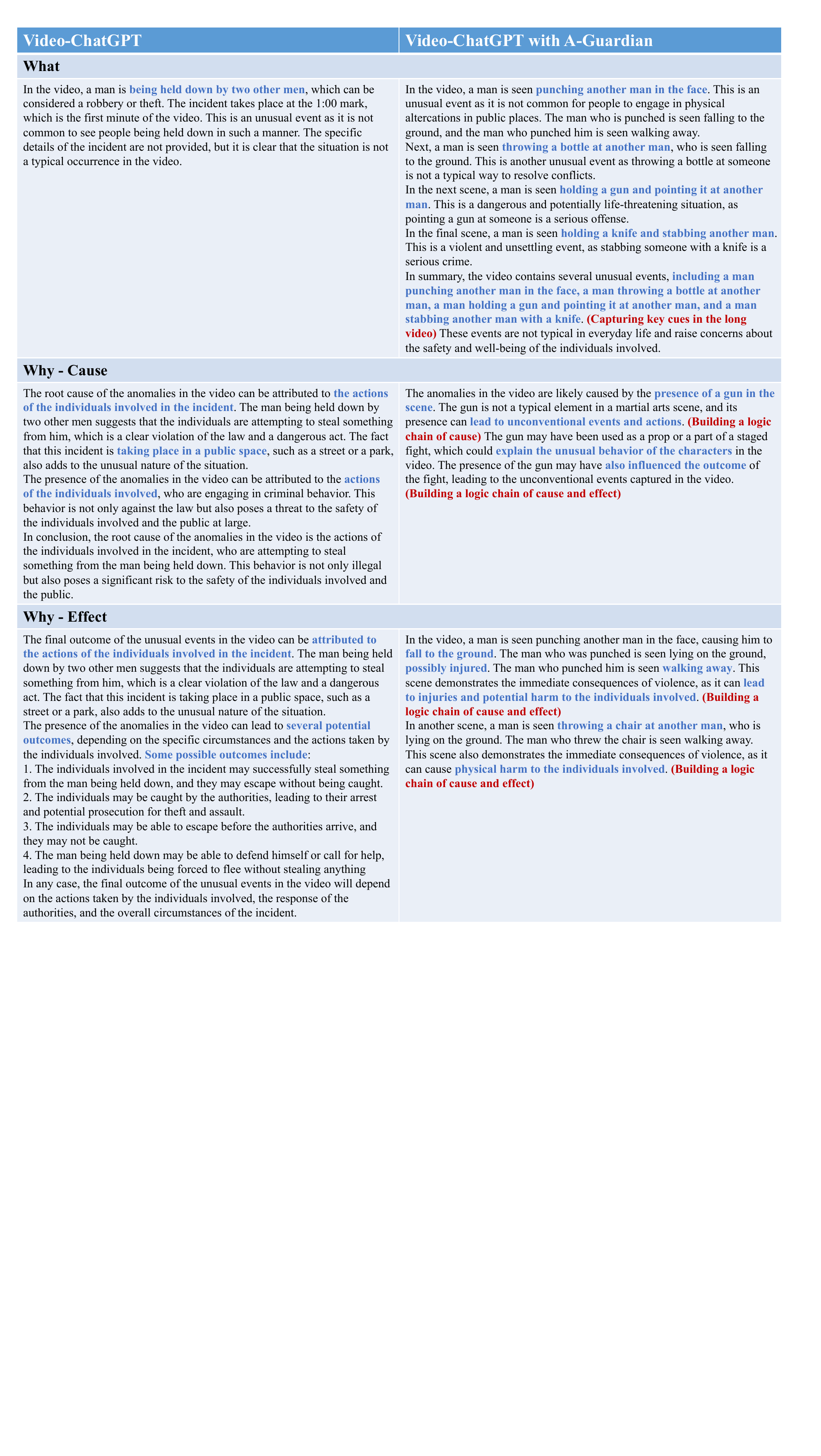}
    \caption{Case study of A-Guardian. We test the Video-ChatGPT model with and without the proposed A-Guardian. Equipped with A-Guardian, Video-ChatGPT can generate a more accurate and detailed description of the anomalies as well as reasonable causes and effects. In the Description task, VLM with A-Guaidian could generate \textit{punching another man in the face} and \textit{holding a gun and pointing it at another man}, whereas Video-ChatGPT without A-Guardian can only provide vague responses about the anomaly event, which proves that the proposed A-Guardian assists the VLM in capturing key cues in the long video. In the Cause task, the model with A-Guardian infers that the gun is the fundamental cause of the anomaly, leading to multiple injuries and people falling.}
    \label{fig:case_study_appendix}
\end{figure*}

We have already shown the case study in section 5.6 of the main paper. 
Here, we further validate the effectiveness of the proposed model A-Guardian by the case study as follows:
Figure \ref{fig:case_study_appendix} shows the results from Video-ChatGPT with and without A-Guardian for the Description, Cause, and Effect tasks related to the same anomalous video. It can be observed that A-Guardian possesses the following abilities:
\begin{itemize}
    \item \textbf{Capturing key cues in the long video:} In the Description task, A-Guardian assists the VLM in identifying crucial moments and events in the anomalous video. For example, \textit{punching another man in the face} and \textit{holding a gun and pointing it at another man}. Whereas Video-ChatGPT without A-Guardian can only provide vague responses about the anomaly event.
    \item \textbf{Building a logic chain of cause and effect:} In the Cause and Effect tasks, A-Guardian guides the model in logical reasoning. In the example, the model with A-Guardian infers that the gun is the fundamental cause of the anomaly, leading to multiple injuries and people falling, followed by the perpetrator leaving. In contrast, the VLM without A-Guardian generates answers that are ambiguous and irrelevant to the events. The result emphasizes the potential results and societal impacts of the anomaly, lacking inference and summarization based on the video content.
 \end{itemize}
\label{appendix c: method}

\end{document}